%% file: main.tex
\documentclass[11pt]{article}

\usepackage{microtype}
\usepackage{graphicx}
\usepackage{subcaption}
\usepackage{booktabs}
\usepackage{microsoft-tech-report}
\usepackage{hyperref}
\usepackage[utf8]{inputenc}
\usepackage{amsmath}
\usepackage{amssymb}
\usepackage{amsfonts}
\usepackage{mathtools}
\usepackage{nicefrac}
\usepackage{enumitem}
\usepackage[capitalize,noabbrev]{cleveref}
\usepackage{xspace}
\usepackage{placeins}
\usepackage{multirow}
\usepackage{pifont}
\usepackage{wrapfig}
\usepackage{soul}
\usepackage{tcolorbox}
\tcbuselibrary{theorems,skins,breakable,listings}

\newtcolorbox{promptbox}[1][]{
  colback=gray!6,
  colframe=gray!40,
  coltitle=black,
  fonttitle=\small\bfseries,
  boxrule=0.5pt,
  arc=2pt,
  left=6pt, right=6pt, top=4pt, bottom=4pt,
  shadow={1pt}{-1pt}{0pt}{gray!30},
  title={#1},
}

\newtcolorbox{findingbox}[1][]{
  enhanced,
  colback=msftcard,
  colframe=msftblue,
  boxrule=0pt,
  leftrule=3pt,
  arc=2pt,
  left=8pt, right=8pt, top=5pt, bottom=5pt,
  before skip=12pt, after skip=12pt,
  fontupper=\normalsize\color{msftdark},
}
\newcommand{\findinglead}[1]{{\msftsans\bfseries\color{msftblue}Finding\ifx\\#1\\\else~(#1)\fi.}\ }

\input{math_commands.tex}

\hypersetup{
  colorlinks=true,
  linkcolor=msftblue,
  citecolor=msftblue,
  urlcolor=msftblue,
  pdftitle={From Raw Experience to Skill Consumption: A Systematic Study of Model-Generated Agent Skills}
}

\techreportshorttitle{SkillLens}

\begin{document}
\thispagestyle{empty}

\noindent
\begin{minipage}[c]{0.5\linewidth}
\raggedright
\raisebox{-0.5\height}{\msftbrandmark}
\end{minipage}
\begin{minipage}[c]{0.49\linewidth}
\raggedleft
{\msftdatefont\small\color{msftgray}May 2026}
\end{minipage}\par
\vspace{0.35em}
\noindent{\color{msftline}\rule{\linewidth}{0.8pt}\par}

\vspace{1.0em}
\begin{center}
{{\msfttitlefont\fontsize{18}{22}\selectfont\color{msftdark}
From Raw Experience to Skill Consumption:\\[0.15em]
A Systematic Study of Model-Generated Agent Skills\par}}
\vspace{1.25em}

{\normalsize\rmfamily\color{msftdark}
Zisu Huang$^{1,2,*,\dagger}$ \hspace{0.6em}
Jingwen Xu$^{1,*}$ \hspace{0.6em}
Yifan Yang$^{2,\ddagger}$ \hspace{0.6em}
Ziyang Gong$^{3}$ \hspace{0.6em}
Qihao Yang$^{3}$\\[0.2em]
Muzhao Tian$^{1}$ \hspace{0.6em}
Xiaohua Wang$^{1}$ \hspace{0.6em}
Changze Lv$^{1}$ \hspace{0.6em}
Xuemei Gao$^{2}$ \hspace{0.6em}
Qi Dai$^{2}$ \hspace{0.6em}
Bei Liu$^{2}$ \hspace{0.6em}
Kai Qiu$^{2}$\\[0.2em]
Xue Yang$^{3}$ \hspace{0.6em}
Dongdong Chen$^{2}$ \hspace{0.6em}
Xiaoqing Zheng$^{1,\ddagger}$ \hspace{0.6em}
Chong Luo$^{2}$\par
}
\vspace{0.22cm}

{\footnotesize\rmfamily\color{msftgray}
$^{1}$ Fudan University \quad
$^{2}$ Microsoft Research \quad
$^{3}$ Shanghai Jiao Tong University\par
}
\end{center}

\vspace{0.45em}
\begin{msfttitlebox}
\setlength{\parindent}{0cm}
\setlength{\parskip}{0.14cm}
\raggedright
\nohyphens

\input{sections/0_abstract}

\vspace{0.14cm}
{\setlength{\parskip}{0.06cm}\small
{\msftmetalabel{Correspondence}
\href{mailto:yifanyang@microsoft.com}{yifanyang@microsoft.com},
\href{mailto:zhengxq@fudan.edu.cn}{zhengxq@fudan.edu.cn}\par}
{\msftmetalabel{Code}\href{https://aka.ms/SkillLens}{https://aka.ms/SkillLens}\par}
}
\vspace{0.08cm}
{\footnotesize\rmfamily\itshape\color{msftgray}
$^*$ Equal contribution. \quad
$^{\dagger}$ Work done during an internship at MSRA. \quad
$^{\ddagger}$ Corresponding authors.\par
}
\end{msfttitlebox}
\suppressfloats[t]

\input{sections/1_introduction}

\input{sections/2_related_work}

\input{sections/3_benchmark}
\input{sections/4_main_results}
\input{sections/5_analysis}

\input{sections/6_quality}
\input{sections/8_conclusion}

\FloatBarrier
\bibliographystyle{unsrtnat}
\bibliography{references}

\clearpage
\input{sections/A_appendix}

\end{document}

%% file: math_commands.tex
\newcommand{\skill}{\mathcal{S}}
\newcommand{\traj}{\mathcal{T}}
\newcommand{\domain}{\mathcal{D}}
\newcommand{\extractor}{E}
\newcommand{\target}{M}
\newcommand{\perf}{\mathrm{Perf}}

\newcommand{\EE}{\mathrm{EE}}

\newcommand{\TE}{\mathrm{TE}}
\newcommand{\sdelta}{\Delta}
\newcommand{\sdeltagap}{\delta}

\newcommand{\better}[1]{\textcolor{green!50!black}{#1}}
\newcommand{\worse}[1]{\textcolor{red!70!black}{#1}}
\definecolor{hlg}{RGB}{200,240,200}
\definecolor{hlr}{RGB}{255,215,215}
\newcommand{\hlgreen}[1]{\sethlcolor{hlg}\hl{#1}}
\newcommand{\hlred}[1]{\sethlcolor{hlr}\hl{#1}}

%% file: sections/0_abstract.tex

Language agents increasingly improve by reusing \emph{skills}---structured procedural artifacts distilled from past experience. In particular, \emph{domain-level} and \emph{model-generated} skills are especially promising. They offer fast adaptation within a domain by encoding domain-specific recurring procedures, and they scale beyond labor-intensive hand-crafting. However, while extraction methods continue to proliferate, understanding remains limited, with no comprehensive study spanning the full skill lifecycle---\textbf{experience generation}, \textbf{skill extraction}, and \textbf{skill consumption}---to ask whether such skills actually work, when they work, and what makes them succeed or fail. To close this gap, we build a utility-grounded evaluation framework that provides systematic experimental results across extractors and target agents, covering five diverse agentic task domains. We find that model-generated skills are beneficial on average but exhibit non-trivial negative transfer, and that neither extractors nor targets behave uniformly. A model can be a strong extractor yet a weak consumer, or vice versa, with skill utility independent of model scale or baseline task strength. To explain these patterns, we then dissect each lifecycle stage in depth, analyzing how experience composition shapes skill quality, what properties characterize useful skills, and how the same skill transfers across different consumers. Finally, we translate these findings into a concrete \emph{meta-skill} that guides skill extraction toward the features tied to actual utility, which consistently improves skill quality across domains and substantially reduces negative transfer.

%% file: sections/1_introduction.tex

\section{Introduction}
\label{sec:intro}

Language agents increasingly improve by reusing knowledge distilled from past trajectories: \emph{skills}---short, structured procedural artifacts---can be loaded at inference time without retraining and have become a defining mechanism for accumulating experience in modern agent stacks~\citep{skillsurvey,agent_survey_luo}.
In particular, \emph{domain-level skills} package a domain's recurring procedures into a single reusable artifact or a coordinated set of them, enabling fast adaptation to new tasks within the domain rather than per-task optimization.
As the practical value of hand-crafted skills has been progressively demonstrated in real-world deployments, skills have become a standard component in several commercial agent platforms~\citep{anthropic_skills}. 
However, hand-crafting skills is labor-intensive and cannot keep pace with the rapidly expanding scope of agent capabilities and deployment.

Therefore, a growing literature turns to \emph{model-generated skills}, producing them automatically at scale~\citep{autorefine,praxis,skillpro,evoskill,skillrl}, with featured works either directly distilling them from execution logs as in Trace2Skill~\citep{trace2skill}, or iteratively refining multi-file skill packages with a co-evolving verifier as in CoEvoSkills~\citep{CoEvoSkills}---offering scalability and automated iteration for agent skills. 
At their core, all these methods follow the same skill lifecycle: generating execution trajectories through agent--environment interaction (\textbf{experience generation}), extracting reusable knowledge or patterns from them (\textbf{skill extraction}), and consuming the resulting skills at inference time (\textbf{skill consumption}).
Despite this methodological momentum, evaluation and understanding lag behind.
Recent benchmarks each illuminate one slice of the picture but leave the lifecycle as a whole opaque.
Most existing efforts study only the skill consumption stage, measuring the marginal performance gain from skill equipment: SkillsBench~\citep{skillsbench} uses task-seeded, human-authored skills, while SWE-Skills-Bench~\citep{sweskillsbench} and Skills-in-the-Wild~\citep{skillwild} draw skills from existing public skill repositories instead---all leaving the skill extraction stage outside the loop.
A notable step toward studying the skill extraction stage is SkillCraft~\citep{skillcraft}, which extracts skills as executable compositions of atomic tools and studies their reuse across tasks. However, it has notable limitations: skills are restricted to executable function compositions, and the benchmark's tasks are designed and scaled to admit such compositions, making it unclear whether the paradigm generalizes to broader domains whose tasks are not designed around function-style reuse.
Taken together, these efforts leave a clear gap: no comprehensive study examines all three stages of the skill lifecycle and systematically asks whether domain-level, model-generated skills actually work, when they work, and what makes them work or fail.

\begin{figure}[t]
    \centering
    \includegraphics[width=\textwidth]{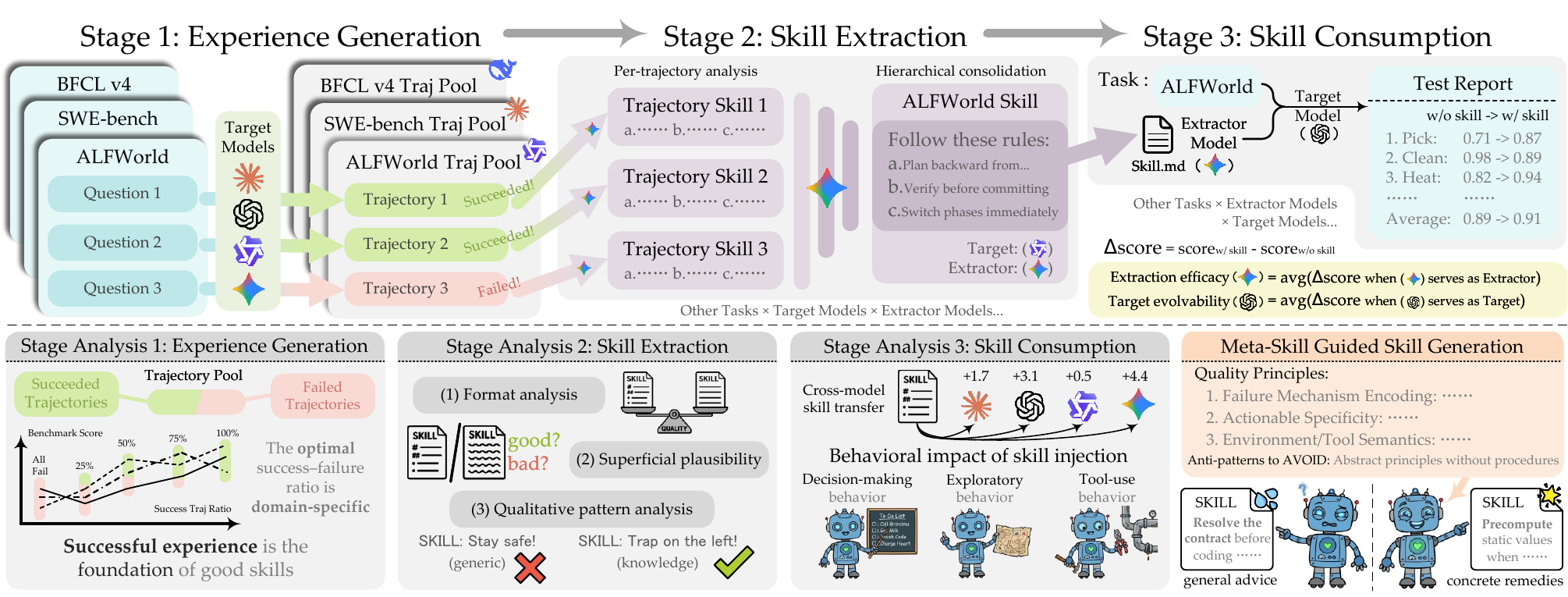}
    \caption{Overview of our study design. We evaluate the full trajectory-to-skill lifecycle across three stages: experience generation, skill extraction, and skill consumption.}
    \label{fig:overview}
\end{figure}

To close this gap, we conduct a comprehensive, utility-grounded study of model-generated, domain-level skills that analyze all three stages of the skill lifecycle.
Specifically, we follow a three-step pipeline: a target agent first executes an experience-generation split to produce an experience pool; an extractor then distills this pool into a single domain-level skill through an extraction framework with minimal design, reflecting the extractor's own ability rather than scaffolding tricks; the resulting skill is finally applied back to the same target and evaluated on the held-out test split to obtain the performance change relative to a no-skill baseline, which we use as a proxy for skill utility.
We instantiate this pipeline across five domains, spanning embodied planning, productivity software, software engineering, web search, and tool calling, and systematically vary the extractor and target.
Based on these experiments, we further introduce two metrics that disentangle the two roles: the \textbf{Extraction Efficacy} ($\EE$)---how reliably a fixed extractor produces helpful skills across targets---and the \textbf{Target Evolvability} ($\TE$)---how much a fixed target benefits from skills extracted by different extractors from its own experience.
Beyond reporting these metrics, we further provide an in-depth analysis spanning all three lifecycle stages, aiming to explain the observed utility patterns and to point toward concrete directions for improving skill extraction.
The pipeline and analysis are summarized in \Cref{fig:overview}.
Overall, our study is organized around three research questions:
\begin{itemize}[leftmargin=*,itemsep=2pt]
    \item \textbf{RQ1}~Do model-generated, domain-level skills reliably benefit downstream agents across targets, extractors, and domains? (Section~\ref{sec:main_results})
    \item \textbf{RQ2}~Across the three lifecycle stages of experience generation (Section~\ref{sec:experience}), skill extraction (Section~\ref{sec:extraction_analysis}), and skill consumption (Section~\ref{sec:consumption}), what actually drives a skill's downstream utility?
    \item \textbf{RQ3}~Can the empirical findings in our study be transformed into a concrete, drop-in improvement to skill extraction itself? (Section~\ref{sec:meta-skill})
\end{itemize}

Answering these questions, we aim to move the entire skill lifecycle from heuristic, intuition-driven practice toward a principled, utility-grounded discipline. As skill libraries proliferate across heterogeneous models and domains, our study helps practitioners obtain skills that are genuinely stable and effective when deployed in real agent systems.

%% file: sections/2_related_work.tex
\section{Related Work}
\label{sec:related}

\paragraph{Automatic Generation of Reusable Knowledge from Agent Experience.}
Recent surveys identify agent skills---composable packages of instructions, code, and resources loaded on demand---as a defining mechanism for extending LLM capabilities without retraining~\citep{skillsurvey}, motivating a growing body of work on automatically extracting such skills from execution trajectories.
These methods scale with collected experience, transfer across tasks and environments, and largely organize around trajectory-to-skill extraction as the core primitive.
\emph{Prompt-based distillation} methods directly summarize trajectories into structured skill artifacts: Trace2Skill~\citep{trace2skill} employs parallel sub-agents followed by hierarchical consolidation, AutoRefine~\citep{autorefine} induces dual-form experience patterns, PRAXIS~\citep{praxis} builds state-indexed procedural memory, and MemP~\citep{memp} formalizes the build--retrieve--update cycle of agent procedural memory.
\emph{Optimization and RL-based methods} further refine extracted skills: ProcMem~\citep{skillpro} applies non-parametric PPO, CoEvoSkills~\citep{CoEvoSkills} uses co-evolutionary verification, and others combine skill banks with reinforcement learning~\citep{evoskill,skillrl,sage}.
A third line studies \emph{self-evolving lifecycle agents} that iteratively refine skills through closed-loop deployment, as in EvolveR~\citep{evolver}.
Despite their differences, all of these approaches rely on trajectory-to-skill extraction as the foundational step that turns raw agent experience into reusable knowledge.
While these works propose effective extraction methods, they each operate under their own setup, and do not provide a systematic understanding spanning the full experience--extraction--consumption lifecycle; our study addresses both gaps through systematic variation across extractors, target models, and domains together with stage-by-stage analysis.

\paragraph{Benchmarks for Agent Skills.}
Recent benchmarks probe complementary aspects of the agent-skill landscape.
One group focuses on \emph{whether skills help at all}: SkillsBench~\citep{skillsbench}, SWE-Skills-Bench~\citep{sweskillsbench}, and \citet{skillwild} primarily test whether curated or discovered skills improve downstream performance over a no-skill baseline.
Another emphasizes \emph{retrieval and orchestration at scale}: AgentSkillOS~\citep{agentskillos} studies ecosystem-level skill management, while SkillFlow~\citep{skillflow} develops scalable retrieval over large skill repositories.
Most closely related to our setting, SkillCraft~\citep{skillcraft} studies \emph{composition and accumulation} via an extraction-and-reuse protocol at test time; however, it restricts skills to executable functions, limiting the diversity of skill representations explored.
Despite this rapid progress, the field still lacks a systematic understanding of the full trajectory-to-skill lifecycle across the raw experience generation, skill extraction, and skill consumption stages.
We address this gap with a comprehensive evaluation framework that crosses skill extractors, skill consumers, and domains, accompanied by detailed analysis of each lifecycle stage.

%% file: sections/3_benchmark.tex
\section{Evaluation Framework}
\label{sec:benchmark}

\subsection{Skill Lifecycle Formulation}
\label{sec:formulation}

Let $\target$ denote a \emph{target model} that both generates experience and consumes skills, and let $\extractor$ denote a (possibly different) \emph{extractor model}. The skill generation lifecycle consists of three stages.

\paragraph{Stage 1: Experience generation.} In domain $\domain$, target model $\target$ executes tasks from the training split $Q^{\text{train}}_\domain$, producing an experience pool $\traj_{\target,\domain}=\{(\text{task}_i,\text{trajectory}_i,\text{outcome}_i)\}$ containing both successful and failed trajectories.

\paragraph{Stage 2: Skill extraction.} $\extractor$ distills the experience pool into a skill set $\skill_{\extractor,\target,\domain}=\extractor(\traj_{\target,\domain})$ using the extraction framework described in Section~\ref{sec:extraction_framework}. The output is structured procedural knowledge under a fixed schema and budget constraint.

\paragraph{Stage 3: Skill consumption.} The same target $\target$ is provided with $\skill_{\extractor,\target,\domain}$ and evaluated on held-out tasks $Q^{\text{test}}_\domain$, measuring how well the extracted skills generalize to unseen tasks in $\domain$.

This protocol simulates a deployment-realistic, extractor-assisted single-step evolution: skills are distilled from $\target$'s own interaction logs and fed back to the same model on held-out tasks, grounding the skill source in $\target$'s actual behavior and failure modes. Holding $\target$ fixed while varying only $\extractor$ enables a controlled comparison of how different extraction procedures convert a model's experience into downstream gains.

\subsection{Extraction Framework}
\label{sec:extraction_framework}

All experiments in our study use a unified extraction framework with intentionally minimal structure: no domain-specific heuristics, filtering rules, or optimization tricks, leaving all abstraction decisions to the extractor model itself. The only imposed organization is a two-stage decomposition that borrows the high-level structure of Trace2Skill~\citep{trace2skill} but strips away its sub-agent fleet, conflict resolution, and skill-deepening mechanisms, retaining only the bare per-trajectory extraction and hierarchical merging steps. This minimal design ensures that performance differences are attributable to extractor capability rather than pipeline engineering.

\paragraph{Per-trajectory analysis.} The extractor $\extractor$ processes each trajectory $\tau_i$ in the experience pool independently, producing a \emph{pattern set} $u_i$ containing multiple success and failure patterns (up to $K$ per trajectory):
\begin{equation}
    \extractor: \tau_i \;\longmapsto\; u_i = \{p_1, \dots, p_k\}, \qquad U = \{u_1, \dots, u_n\}
    \label{eq:map}
\end{equation}
Each pattern captures a reusable behavioral insight: \emph{success patterns} encode strategies that led to task completion, while \emph{failure patterns} encode error modes and pitfalls. Since trajectories are processed independently, this phase is fully parallelizable.

\paragraph{Hierarchical consolidation.} The extractor $\extractor$ then consolidates the pattern sets in a tree-structured reduction with configurable group size $G$: at each level, $\extractor$ merges $G$ pattern sets by deduplicating, generalizing, and reconciling overlapping patterns until a single consolidated pattern set remains:
\begin{equation}
    U^{(0)} = U, \qquad U^{(\ell+1)} = \bigl\{\textsc{Merge}_{\extractor}\bigl(u^{(\ell)}_{G(j{-}1)+1},\, \dots,\, u^{(\ell)}_{Gj}\bigr)\bigr\}_{j}, \qquad \text{until } |U^{(L)}| = 1
    \label{eq:reduce}
\end{equation}
Finally, $\extractor$ converts the consolidated pattern set into the skill set $\skill_{\extractor,\target,\domain}$ via structured tool-calling operations that support creation, update, and deletion of skills with schema validation.

\paragraph{Skill representation.} Each skill follows a fixed schema based on the Agent Skills open standard\footnote{\url{https://github.com/agentskills/agentskills}}, with fields for \texttt{name}, \texttt{description}, \texttt{body} (Markdown procedural instructions), and optional \texttt{references} and \texttt{scripts}.

\subsection{Evaluation Metric}
\label{sec:metric}

We evaluate the effectiveness of extracted skills by downstream performance gain rather than text quality. For each extractor--target--domain triple $(\extractor,\target,\domain)$, we measure the performance delta caused by injecting the extracted skill:
\begin{equation}
    \sdelta(\extractor, \target, \domain) \;=\; \perf(\target \mid \skill_{\extractor,\target,\domain},\; Q^{\text{test}}_\domain) \;-\; \perf(\target \mid Q^{\text{test}}_\domain)
    \label{eq:delta}
\end{equation}
where $\perf$ is the domain-specific task metric. Baseline and skill-augmented evaluations use the same held-out split $Q^{\text{test}}_\domain$. $\sdelta>0$ indicates improvement and $\sdelta<0$ indicates negative transfer.

For each domain, varying $\extractor$ and $\target$ yields the set $\{\sdelta(\extractor,\target,\domain): \extractor \in \mathcal{E},\target \in \mathcal{M}\}$, where $\mathcal{E}$ is the set of extractors and $\mathcal{M}$ is the set of target models. We summarize these extractor--target performance gains from two complementary perspectives for deeper insights:

\paragraph{Extraction efficacy.}
This metric captures the extractor-side effect. For a fixed extractor, it asks how reliably that extractor converts different target-specific experience pools into skills that improve downstream performance:
\begin{equation}
    \EE(\extractor, \domain) = \frac{1}{|\mathcal{M}|} \sum_{\target \in \mathcal{M}} \sdelta(\extractor, \target, \domain).
    \label{eq:ee}
\end{equation}

\paragraph{Target evolvability.}
This metric captures the target-side effect. For a fixed target, it asks how much the target improves when different extractors distill skills from the target's own experience and feed them back to the same target:
\begin{equation}
    \TE(\target, \domain) = \frac{1}{|\mathcal{E}|} \sum_{\extractor \in \mathcal{E}} \sdelta(\extractor, \target, \domain).
    \label{eq:te}
\end{equation}
We report both $\EE$ and $\TE$ per domain, since task metrics and difficulty are domain-specific. We also retain each extractor--target $\sdelta$ to analyze interactions beyond these averages.

%% file: sections/4_main_results.tex
\section{Main Experiments}
\label{sec:main_results}

\newcommand{\modelicon}[1]{\raisebox{-0.12em}{\makebox[0.95em][c]{\includegraphics[height=0.82em]{#1}}}}
\newcommand{\gptmodel}[1]{\modelicon{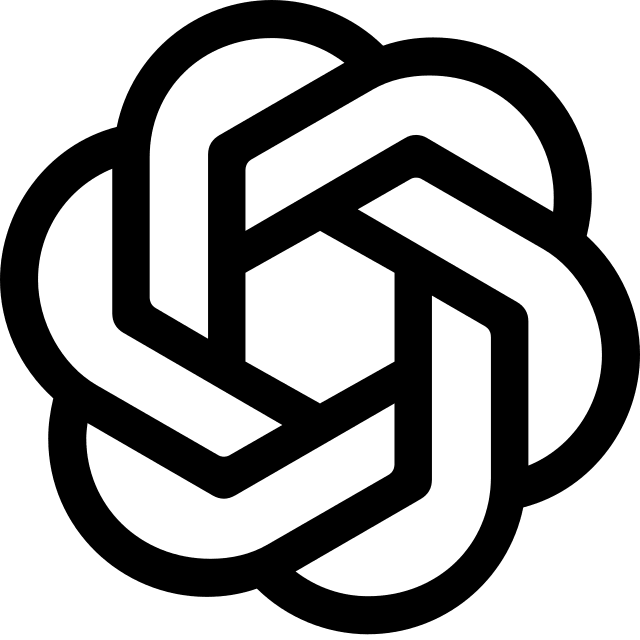}\,#1}
\newcommand{\geminimodel}[1]{\modelicon{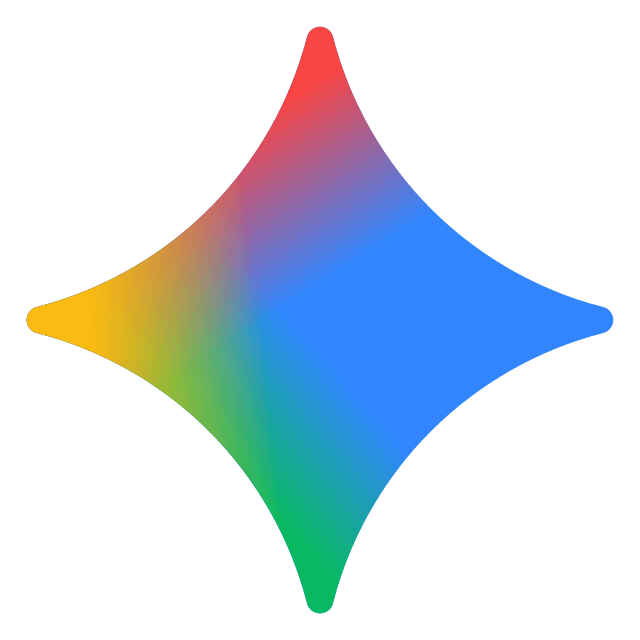}\,#1}
\newcommand{\qwenmodel}[1]{\modelicon{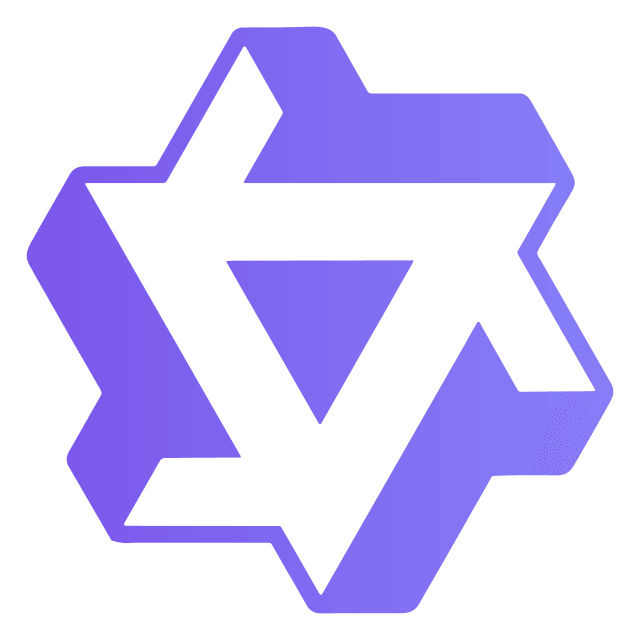}\,#1}

\begin{table*}[!t]
    \centering
    \small
    \renewcommand{\arraystretch}{0.92}
    \setlength{\tabcolsep}{9.3pt}
    \begin{tabular}{@{\hspace{2pt}}lcccccc@{\hspace{8pt}}|@{\hspace{8pt}}c@{\hspace{6pt}}}
        \toprule
        Target & Base & \gptmodel{5.4} & \gptmodel{5.4-mini} & \geminimodel{3.1-Pro} & \geminimodel{3.1-FL} & \qwenmodel{3.5-35B} & $\TE$ \\
        \midrule
        \rowcolor{blue!8} \multicolumn{8}{c}{Embodied: \textit{ALFWorld}} \\[2pt]
        \gptmodel{5.4}      & 68.66 & \textcolor{green!50!black}{+1.49} & \textcolor{green!50!black}{+6.47} & \textcolor{green!50!black}{+7.46} & \textcolor{green!50!black}{+4.98} & \textcolor{green!50!black}{+4.23} & \textbf{+4.93} \\
        \gptmodel{5.4-mini} & 52.24 & \textcolor{green!50!black}{+1.00} & \textcolor{green!50!black}{+4.23} & \textcolor{green!50!black}{+2.74} & \textcolor{green!50!black}{+2.24} & \textcolor{green!50!black}{+3.98} & \underline{+2.84} \\
        \geminimodel{3.1-Pro}  & 87.56 & \textcolor{green!50!black}{+0.50} & \textcolor{green!50!black}{+0.75} & \textcolor{green!50!black}{+0.00} & \textcolor{red!70!black}{$-$0.75} & \textcolor{red!70!black}{$-$1.24} & $-$0.15 \\
        \geminimodel{3.1-FL}   & 51.99 & \textcolor{red!70!black}{$-$2.49} & \textcolor{red!70!black}{$-$1.24} & \textcolor{green!50!black}{+1.49} & \textcolor{red!70!black}{$-$2.49} & \textcolor{red!70!black}{$-$3.23} & $-$1.59 \\
        \qwenmodel{3.5-35B}  & 57.21 & \textcolor{red!70!black}{$-$1.99} & \textcolor{red!70!black}{$-$3.48} & \textcolor{red!70!black}{$-$0.75} & \textcolor{green!50!black}{+0.50} & \textcolor{red!70!black}{$-$1.00} & $-$1.34 \\
        \qwenmodel{3.5-9B}   & 36.07 & \textcolor{red!70!black}{$-$2.49} & \textcolor{red!70!black}{$-$2.99} & \textcolor{red!70!black}{$-$1.24} & \textcolor{red!70!black}{$-$1.99} & \textcolor{green!50!black}{+0.25} & $-$1.69 \\
        \cmidrule{1-8}
        \rowcolor{gray!5} $\EE$ & & $-$0.66 & \underline{+0.62} & \textbf{+1.62} & +0.42 & +0.50 & \\
        \midrule
        \rowcolor{blue!8} \multicolumn{8}{c}{Productivity: \textit{SpreadsheetBench}} \\[2pt]
        \gptmodel{5.4}      & 37.17 & \textcolor{green!50!black}{+4.33} & \textcolor{green!50!black}{+9.00} & \textcolor{green!50!black}{+14.00} & \textcolor{green!50!black}{+14.66} & \textcolor{green!50!black}{+6.33} & \textbf{+9.66} \\
        \gptmodel{5.4-mini} & 29.33 & \textcolor{green!50!black}{+0.34} & \textcolor{green!50!black}{+2.50} & \textcolor{green!50!black}{+3.67} & \textcolor{green!50!black}{+4.50} & \textcolor{green!50!black}{+1.00} & +2.40 \\
        \geminimodel{3.1-Pro}  & 35.83 & \textcolor{red!70!black}{$-$0.50} & \textcolor{red!70!black}{$-$2.67} & \textcolor{green!50!black}{+6.50} & \textcolor{green!50!black}{+5.33} & \textcolor{green!50!black}{+5.83} & +2.90 \\
        \geminimodel{3.1-FL}   & 25.00 & \textcolor{green!50!black}{+2.67} & \textcolor{green!50!black}{+1.83} & \textcolor{green!50!black}{+1.50} & \textcolor{green!50!black}{+6.17} & \textcolor{green!50!black}{+7.33} & \underline{+3.90} \\
        \qwenmodel{3.5-35B}  & 23.83 & \textcolor{green!50!black}{+2.00} & \textcolor{green!50!black}{+5.50} & \textcolor{green!50!black}{+0.17} & \textcolor{green!50!black}{+3.34} & \textcolor{red!70!black}{$-$3.50} & +1.50 \\
        \qwenmodel{3.5-9B}   & 13.67 & \textcolor{green!50!black}{+1.16} & \textcolor{green!50!black}{+3.16} & \textcolor{red!70!black}{$-$1.17} & \textcolor{green!50!black}{+1.16} & \textcolor{green!50!black}{+3.00} & +1.46 \\
        \cmidrule{1-8}
        \rowcolor{gray!5} $\EE$ & & +1.67 & +3.22 & \underline{+4.11} & \textbf{+5.86} & +3.33 & \\
        \midrule
        \rowcolor{blue!8} \multicolumn{8}{c}{Coding: \textit{SWE-bench-Verified}} \\[2pt]
        \gptmodel{5.4}      & 68.40 & \textcolor{green!50!black}{+4.67} & \textcolor{green!50!black}{+1.33} & \textcolor{green!50!black}{+2.00} & \textcolor{green!50!black}{+4.00} & \textcolor{green!50!black}{+2.27} & \underline{+2.85} \\
        \gptmodel{5.4-mini} & 59.73 & \textcolor{green!50!black}{+3.20} & \textcolor{green!50!black}{+3.20} & \textcolor{green!50!black}{+1.73} & \textcolor{green!50!black}{+3.60} & \textcolor{green!50!black}{+2.80} & \textbf{+2.91} \\
        \geminimodel{3.1-Pro}  & 66.53 & \textcolor{green!50!black}{+2.00} & \textcolor{green!50!black}{+2.80} & \textcolor{green!50!black}{+2.13} & \textcolor{green!50!black}{+3.47} & \textcolor{red!70!black}{$-$1.60} & +1.76 \\
        \geminimodel{3.1-FL}   & 55.47 & \textcolor{green!50!black}{+2.67} & \textcolor{green!50!black}{+3.33} & \textcolor{green!50!black}{+2.93} & \textcolor{green!50!black}{+3.47} & \textcolor{red!70!black}{$-$0.93} & +2.29 \\
        \qwenmodel{3.5-35B}  & 52.93 & \textcolor{green!50!black}{+3.20} & \textcolor{green!50!black}{+2.00} & \textcolor{green!50!black}{+2.53} & \textcolor{green!50!black}{+2.93} & \textcolor{green!50!black}{+2.00} & +2.53 \\
        \qwenmodel{3.5-9B}   & 33.33 & \textcolor{red!70!black}{$-$1.07} & \textcolor{green!50!black}{+2.40} & \textcolor{red!70!black}{$-$1.60} & \textcolor{green!50!black}{+1.20} & \textcolor{green!50!black}{+0.93} & +0.37 \\
        \cmidrule{1-8}
        \rowcolor{gray!5} $\EE$ & & +2.45 & \underline{+2.51} & +1.62 & \textbf{+3.11} & +0.91 & \\
        \midrule
        \rowcolor{blue!8} \multicolumn{8}{c}{Web Search: \textit{SEAL-0}} \\[2pt]
        \gptmodel{5.4}      & 51.24 & \textcolor{green!50!black}{+6.47} & \textcolor{green!50!black}{+4.23} & \textcolor{green!50!black}{+7.71} & \textcolor{green!50!black}{+1.74} & \textcolor{green!50!black}{+1.74} & \underline{+4.38} \\
        \gptmodel{5.4-mini} & 45.27 & \textcolor{red!70!black}{$-$1.49} & \textcolor{green!50!black}{+3.23} & \textcolor{red!70!black}{$-$3.98} & \textcolor{green!50!black}{+3.98} & \textcolor{red!70!black}{$-$4.23} & $-$0.50 \\
        \geminimodel{3.1-Pro}  & 55.97 & \textcolor{red!70!black}{$-$4.23} & \textcolor{red!70!black}{$-$1.99} & \textcolor{green!50!black}{+1.99} & \textcolor{green!50!black}{+2.49} & \textcolor{red!70!black}{$-$3.48} & $-$1.04 \\
        \geminimodel{3.1-FL}   & 14.93 & \textcolor{green!50!black}{+9.45} & \textcolor{green!50!black}{+8.21} & \textcolor{green!50!black}{+2.99} & \textcolor{red!70!black}{$-$1.24} & \textcolor{green!50!black}{+7.21} & \textbf{+5.32} \\
        \qwenmodel{3.5-35B}  & 40.55 & \textcolor{green!50!black}{+1.74} & \textcolor{green!50!black}{+6.47} & \textcolor{red!70!black}{$-$3.73} & \textcolor{green!50!black}{+4.73} & \textcolor{green!50!black}{+2.24} & +2.29 \\
        \qwenmodel{3.5-9B}   & 33.83 & \textcolor{green!50!black}{+10.70} & \textcolor{green!50!black}{+8.96} & \textcolor{red!70!black}{$-$5.72} & \textcolor{green!50!black}{+5.97} & \textcolor{red!70!black}{$-$2.99} & +3.38 \\
        \cmidrule{1-8}
        \rowcolor{gray!5} $\EE$ & & \underline{+3.77} & \textbf{+4.85} & $-$0.12 & +2.95 & +0.08 & \\
        \midrule
        \rowcolor{blue!8} \multicolumn{8}{c}{Tool Calling: \textit{BFCL-v4}} \\[2pt]
        \gptmodel{5.4}      & 51.68 & \textcolor{green!50!black}{+3.08} & \textcolor{green!50!black}{+5.04} & \textcolor{green!50!black}{+0.42} & \textcolor{green!50!black}{+5.04} & \textcolor{red!70!black}{$-$2.24} & +2.27 \\
        \gptmodel{5.4-mini} & 53.50 & \textcolor{green!50!black}{+3.92} & \textcolor{green!50!black}{+6.16} & \textcolor{green!50!black}{+7.56} & \textcolor{green!50!black}{+5.18} & \textcolor{green!50!black}{+2.94} & \textbf{+5.15} \\
        \geminimodel{3.1-Pro}  & 51.82 & \textcolor{green!50!black}{+5.32} & \textcolor{green!50!black}{+5.88} & \textcolor{red!70!black}{$-$4.34} & \textcolor{green!50!black}{+0.14} & \textcolor{green!50!black}{+6.02} & +2.60 \\
        \geminimodel{3.1-FL}   & 41.18 & \textcolor{green!50!black}{+4.06} & \textcolor{green!50!black}{+4.62} & \textcolor{green!50!black}{+4.20} & \textcolor{green!50!black}{+8.12} & \textcolor{green!50!black}{+3.64} & \underline{+4.93} \\
        \qwenmodel{3.5-35B}  & 57.56 & \textcolor{red!70!black}{$-$0.70} & \textcolor{green!50!black}{+0.84} & \textcolor{green!50!black}{+1.54} & \textcolor{green!50!black}{+2.80} & \textcolor{green!50!black}{+1.82} & +1.26 \\
        \qwenmodel{3.5-9B}   & 46.78 & \textcolor{green!50!black}{+2.94} & \textcolor{green!50!black}{+2.94} & \textcolor{green!50!black}{+1.82} & \textcolor{red!70!black}{$-$0.98} & \textcolor{red!70!black}{$-$1.12} & +1.12 \\
        \cmidrule{1-8}
        \rowcolor{gray!5} $\EE$ & & +3.10 & \textbf{+4.25} & +1.87 & \underline{+3.38} & +1.84 & \\
        \bottomrule
    \end{tabular}
    \caption{Skill-induced performance gain ($\sdelta$) across domains. \textit{Base} is the no-skill baseline. \textbf{$\TE$} denotes \textbf{Target Evolvability}, averaged across extractors, and \textbf{$\EE$} denotes \textbf{Extraction Efficacy}, averaged across targets. \textcolor{green!50!black}{Green}: $\sdelta > 0$; \textcolor{red!70!black}{Red}: $\sdelta < 0$.}
    \label{tab:cross_matrix}
\end{table*}

In this section, we conduct a large-scale evaluation of model-generated agent skills across five domains, six target models, and five extractor models. The goal is to characterize when extracted skills improve downstream performance, when they fail or degrade it, and how these outcomes vary across the extractor--target--domain space. We report the main empirical patterns here and leave deeper analysis to Section~\ref{sec:analysis}.

\subsection{Experimental Setup}
\label{sec:setup}

\paragraph{Domains.} To obtain a comprehensive view of model-generated skills, our evaluation spans five qualitatively different domains: embodied interaction, productivity software, software engineering, web search, and tool calling. This breadth lets us test whether extracted skills remain useful across different forms of agent behavior:
\begin{itemize}[leftmargin=*,itemsep=2pt]
    \item \textbf{ALFWorld}~\citep{alfworld}: embodied household tasks requiring physical commonsense, exploration, and multi-step planning.
    \item \textbf{SpreadsheetBench}~\citep{spreadsheetbench}: spreadsheet manipulation tasks involving table inspection, formula reasoning, filtering, and value editing.
    \item \textbf{SWE-bench-Verified}~\citep{swebench}: real-world software engineering tasks requiring codebase understanding, fault localization, and patch generation.
    \item \textbf{SEAL-0}~\citep{sealqa}: web-search question answering tasks requiring retrieval, evidence synthesis, and multi-hop reasoning.
    \item \textbf{BFCL-v4}~\citep{bfcl}: tool-calling tasks requiring function selection, parameter extraction, type matching, and multi-turn tool use. We use the \emph{multi-turn} subset, which exercises long-horizon, procedural tool-use behaviour relevant to skill reuse.
\end{itemize}

\paragraph{Models.} We select models spanning different families and scales: \textbf{GPT} (GPT-5.4, GPT-5.4-mini)~\citep{openai2026gpt54}, \textbf{Gemini} (Gemini-3.1-Pro~\citep{google2026gemini31pro}, Gemini-3.1-Flash-Lite~\citep{google2026gemini31flashlite}), and \textbf{Qwen} (Qwen3.5-35B, Qwen3.5-9B)~\citep{qwen3.5}. All six models serve as targets. During preliminary experiments, we found that Qwen3.5-9B cannot reliably follow the structured extraction protocol (Section~\ref{sec:extraction_framework}), so it is excluded as an extractor.

\paragraph{Data splits and evaluation protocol.} For each domain $\domain$, we split task instances 1:1 into a experience-generation split $Q^{\text{train}}_\domain$ and a held-out test split $Q^{\text{test}}_\domain$; if an official training split exists, $Q^{\text{train}}_\domain$ is sampled from it at the same proportion. Each target $\target$ runs $Q^{\text{train}}_\domain$ to form a experience pool $\traj_{\target,\domain}$. Each extractor $\extractor$ distills this pool into a single consolidated skill $\skill_{\extractor,\target,\domain}$ in our main experiments, which is supplied in the target's system prompt at inference time and evaluated on $Q^{\text{test}}_\domain$. We run each evaluation three times and report the average $\sdelta$ (Eq.~\ref{eq:delta}) in percentage points. Full extraction and evaluation details are in Appendix~\ref{app:details}.

\subsection{Main Results}
\label{sec:target_dependent}

The following results answer \textbf{RQ1}: whether model-generated, domain-level skills reliably benefit downstream agents across targets, extractors, and domains. We report per-cell performance deltas across the extractor--target matrix together with the aggregated $\EE$ and $\TE$ metrics.

\paragraph{Model-generated skills are generally beneficial, but not guaranteed.} Table~\ref{tab:cross_matrix} presents the full $\sdelta$ matrix across domains. Model-generated skills are generally effective, improving downstream performance in 75\% of entries. Yet negative transfer remains common: 25\% of entries have $\sdelta < 0$, meaning that applying extracted skills degrades the target's performance. This risk is domain-dependent: SpreadsheetBench and SWE-bench-Verified have the lowest negative rates (13\%), whereas ALFWorld is the most fragile domain (47\%). Thus, positive average gains mask a substantial risk of negative transfer, so model-generated skills cannot be assumed to improve performance.

\paragraph{Better executor is not necessarily better extractor.}
Extractor-side performance does not simply follow model scale or baseline task strength. For example, on SpreadsheetBench, the lightweight Gemini-3.1-Flash-Lite achieves the highest $\EE$, while GPT-5.4 ranks last despite having the strongest baseline among the targets. This reversal shows that skill extraction is a distinct capability from task execution: the extractor must convert target-specific trajectories into procedural guidance that the target can actually exploit. Consequently, choosing an extractor is not equivalent to choosing the strongest model; it is a compatibility problem between extractor, target, and domain.

\paragraph{Skill utility is target-dependent.}
Even within the same domain, the same set of extractors can produce very different gains across targets. On ALFWorld, GPT-5.4 benefits consistently from all five extractors ($\TE = +4.93$), while Gemini-3.1-Flash-Lite, Qwen3.5-35B, and Qwen3.5-9B all have negative $\TE$. Similar asymmetries appear across other domains. This suggests that skill benefit is shaped not only by extractor quality, but also by what a target's own experience makes extractable and what the target can execute from the resulting guidance.

\begin{findingbox}
\findinglead{RQ1}Model-generated, domain-level skills help on average but are \emph{not} universally beneficial: 25\% of evaluated extractor--target pairings exhibit negative transfer. An extractor's task-solving capability does not predict its extraction quality, and $\TE$ varies sharply across targets even within a single domain---both the extractor and the target jointly shape skill utility.
\end{findingbox}

%% file: sections/5_analysis.tex
\section{Diving Deeper into the Agent Skill Lifecycle}
\label{sec:analysis}

This section addresses \textbf{RQ2}: \emph{what actually drives a skill's downstream utility?} Following the lifecycle defined in \Cref{fig:overview}, we further analyze the three stages separately---\textbf{experience generation}, \textbf{skill extraction}, and \textbf{skill consumption}, and ask what factors at each stage govern downstream gains.

\subsection{Experience Generation: Success or Failure, Which Teaches Better Skills?}
\label{sec:experience}

The first stage determines what information is available for extraction. A natural and key factor is the success/failure composition of the experience pool: successful trajectories expose workable procedures, while failures may expose constraints and pitfalls. We isolate this factor by directly manipulating pool composition.

\paragraph{Setup.}
We fix the extractor (GPT-5.4-mini) and sample five experience pools from the same source trajectories, with success ratios of 100\%, 75\%, 50\%, 25\%, and 0\%. Each pool is converted into a skill using the same extraction pipeline. We evaluate the resulting skills on SpreadsheetBench, SWE-bench-Verified, and ALFWorld with three targets and report the average $\sdelta$.

\begin{wrapfigure}{r}{0.53\textwidth}
\centering
\includegraphics[width=0.53\textwidth]{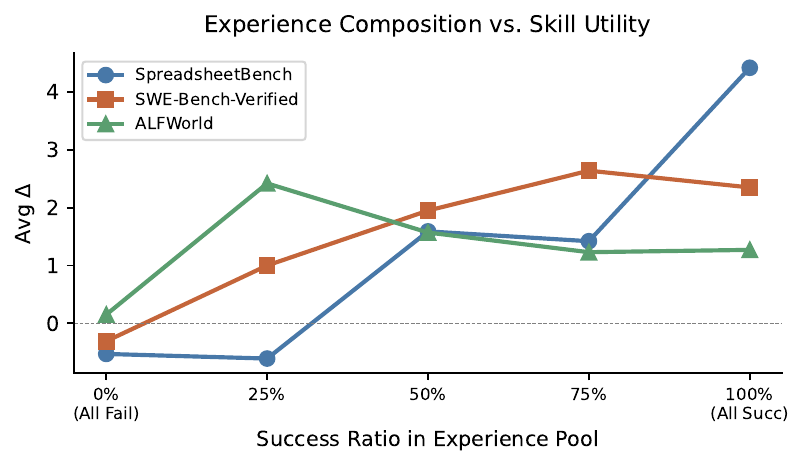}
\caption{Effect of success ratio in the experience pool on downstream tasks.}
\label{fig:experience_ratio}
\end{wrapfigure}

\paragraph{Results.} \Cref{fig:experience_ratio} shows that \textbf{experience composition strongly affects extracted skill quality}.
Beyond this, \textbf{the optimal success--failure ratio is domain-specific}. SpreadsheetBench favors more successful trajectories, SWE-bench-Verified peaks with a mostly successful pool, and ALFWorld performs best with failure-heavy pools. This suggests that domain-specific behavior patterns shape the informational value of successes versus failures for skill extraction: in ALFWorld, for example, failed attempts often reveal invalid actions and dead-end states, making failures surprisingly informative.
Overall, Figure~\ref{fig:experience_ratio} also shows that all-failure pools consistently perform worst, highlighting \textbf{successful trajectories as the foundation of skill extraction}: they provide positive procedural signals that guide the agent's actions and narrow its exploration space, rather than merely indicating what to avoid.

\begin{findingbox}
\findinglead{Experience}Skill quality appears to be largely influenced by the success--failure composition of the experience pool. Pure-failure pools are consistently the worst, while the optimal mix tends to be domain-dependent---reflecting how each domain weighs positive procedural signals against negative constraint signals.
\end{findingbox}

\subsection{Skill Extraction: What Makes a Good Skill?}
\label{sec:extraction_analysis}

Given that experience quality matters (Section~\ref{sec:experience}), we now ask whether shallow textual features of a skill can explain its downstream gains. We rule out two such candidates and surface a qualitative pattern that motivates the systematic analysis in Section~\ref{sec:meta-skill}.

\paragraph{Skill quality is not reducible to surface form.}
A natural first concern is that skill format may largely influence skill utility.
We test this by rewriting the same skill into four canonical formats (\emph{ordered list}, \emph{unordered list}, \emph{checklist}, and \emph{prose}) and re-evaluating each rewrite.
We then run a Friedman test, which ranks the four formats within each task and asks whether some format is consistently ranked higher than the others across tasks.
Results in \Cref{tab:format_norm} (Appendix~\ref{app:format}) show that the format effect is non-significant on every target (all $p{>}0.34$), whereas swapping the extractor produces a clearly discernible effect on 5/6 targets ($p{<}0.01$).
This contrast indicates that variance is driven by what a skill says, not how it looks.

\begin{wrapfigure}{r}{0.5\textwidth}
\centering
\includegraphics[width=0.48\textwidth]{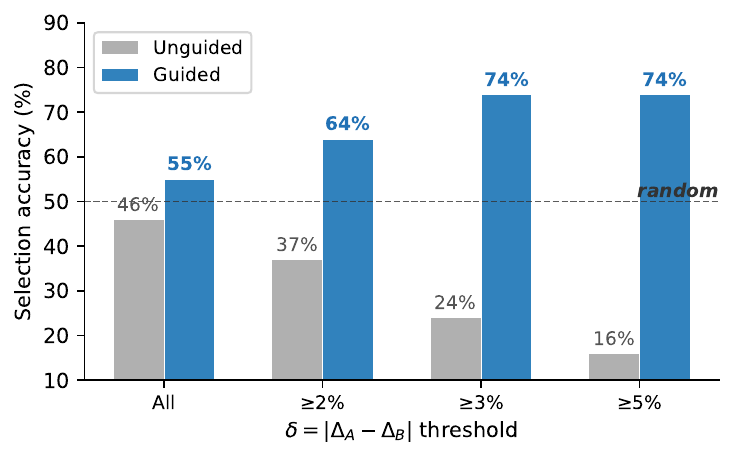}
\caption{Pairwise selection accuracy by $\sdeltagap$.}
\label{fig:pairwise}
\end{wrapfigure}

\paragraph{Textual plausibility does not predict skill utility.}
If content matters, can we identify better skills from the text alone?
We probe this with a GPT-5.4 judge as a human proxy. For a pair of skills extracted within the same $(\target,\domain)$, the judge sees only the two skill texts and selects the one it deems higher-quality (better downstream performance).
We evaluate on 151 pairs whose $\sdeltagap=|\sdelta_A - \sdelta_B|$ exceeds 0.5\%, excluding near-ties (details in Appendix~\ref{app:format}).
Without any evaluation criteria, overall LLM selection accuracy is 46.4\%, indistinguishable from random.

The gray bars in \Cref{fig:pairwise} break this number down by $\sdeltagap$: more strikingly, accuracy \emph{decreases} as $\sdeltagap$ grows. On pairs with $\sdeltagap{\geq}5\%$, the judge picks the higher-$\sdelta$ skill only 15.8\% of the time, a clear inversion of actual utility.
In other words, the skill that reads better is often the one that performs worse. Textual plausibility has come apart from downstream skill utility, a gap we close in Section~\ref{sec:meta-skill} by identifying which textual properties carry genuine predictive signal.

\paragraph{A qualitative hint: concrete remedies, not generic advice.}
A qualitative inspection of one high-$\sdeltagap$ pair in SpreadsheetBench (\Cref{tab:case_ss}, Appendix~\ref{app:contrastive}) hints at where the gap lies. The higher-$\sdelta$ skill names concrete failure mechanisms with executable remedies (e.g., precomputing static values when host engines do not evaluate formula strings), whereas the lower-$\sdelta$ skill offers only generic procedural advice (e.g., ``resolve the contract before coding''). We treat this as motivation; Section~\ref{sec:meta-skill} tests at scale which textual dimensions actually predict utility.

\begin{findingbox}
\findinglead{Extraction}Neither skill format nor textual plausibility predicts utility: directly asking an LLM to judge the skill text performs no better than chance. What truly drives skill utility lies deeper than surface form, motivating the meta-skill analysis in Section~\ref{sec:meta-skill}.
\end{findingbox}

\subsection{Skill Consumption: How Does Skill Benefit Vary Across Target Models?}
\label{sec:consumption}

Sections~\ref{sec:experience}--\ref{sec:extraction_analysis} focus on the \emph{supply side} of skills: what experience to extract from and what textual properties matter. Here we turn to the \emph{demand side}: given an identical skill, how much benefit does each consumer actually derive?

\begin{wrapfigure}{r}{0.55\textwidth}
\centering
\includegraphics[width=0.53\textwidth]{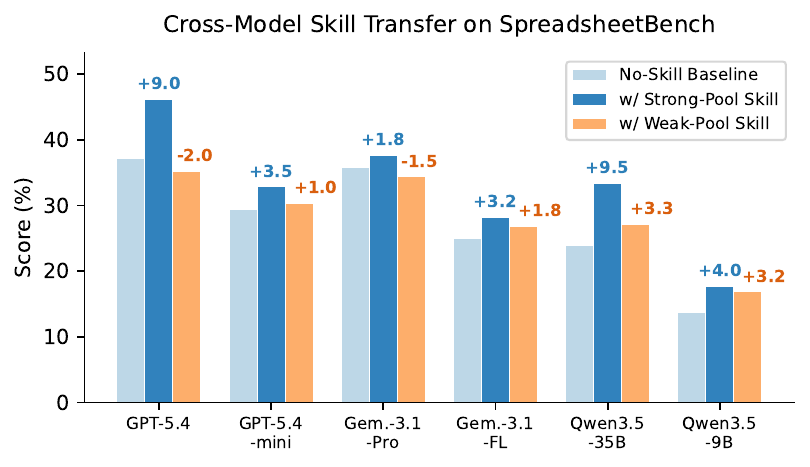}
\caption{Cross-model skill transfer. Strong-pool and weak-pool skills are injected into each target separately.}
\label{fig:consumption_transfer}
\end{wrapfigure}

\paragraph{Cross-model skill transfer.}
Our goal here is to ask how the same skill behaves when consumed by different targets. To sharpen the comparison, we fix a single extractor (GPT-5.4-mini) and select two contrasting skills from its main-experiment outputs on SpreadsheetBench: a \emph{strong-pool skill} distilled from the strongest baseline target's experience pool (GPT-5.4) and a \emph{weak-pool skill} from the weakest (Qwen3.5-9B). These two skills are applied to all six targets. Two patterns emerge in the results shown in \Cref{fig:consumption_transfer}. First, with the skill text held fixed, per-target gains differ sharply---the strong-pool skill ranges from $+1.8$ on Gem-3.1-Pro to $+9.5$ on Qwen3.5-35B, and a similar spread holds for the weak-pool skill---showing that \textbf{skill consumption ability varies across targets}. Second, the strong-pool skill consistently improves every target, whereas the weak-pool skill yields clear negative transfer on some targets (e.g., $-2.0$ on GPT-5.4) and only modest gains on others---a gap that, in turn, echoes the Section~\ref{sec:experience} finding that \textbf{experience-pool quality is critical to the skills it produces}.

\paragraph{Behavioral impact of skill consumption.}
To understand why skill consumption helps some targets but hurts others, we systematically examine agent trajectories on two contrasting targets on SpreadsheetBench: GPT-5.4, which improves substantially after consuming skills, and Qwen3.5-9B, which regresses under certain skills. We characterize the observed changes along three axes: \textbf{decision-making behavior}, i.e., what solution strategy the model chooses; \textbf{exploratory behavior}, i.e., how the agent builds an understanding of the workbook and task environment before acting; and \textbf{tool-use behavior}, i.e., how the chosen strategy is instantiated through concrete operations (details in Appendix~\ref{app:behavior}). Across both targets, skill consumption \emph{reshapes the default policy} rather than triggering new explicit skill calls: it steers GPT-5.4 toward evaluator-aligned computation and verification, while pushing Qwen3.5-9B toward complex workbook-native workflows that gain structural fidelity at the cost of execution robustness.

\begin{findingbox}
\findinglead{Consumption}Given the \emph{same} skill text, per-target gains can differ substantially, with some targets benefiting strongly while others see little effect or even regress. Skill consumption acts by reshaping the target's default policy, so consumption ability is itself a per-target property that bounds achievable gains.
\end{findingbox}

%% file: sections/6_quality.tex
\section{From Diagnosis to Intervention: Meta-Skill Guided Extraction}
\label{sec:meta-skill}

\begin{figure}[t]
\centering
\hspace*{-0.025\linewidth}\includegraphics[width=0.9\linewidth,trim=4pt 4pt 4pt 4pt,clip]{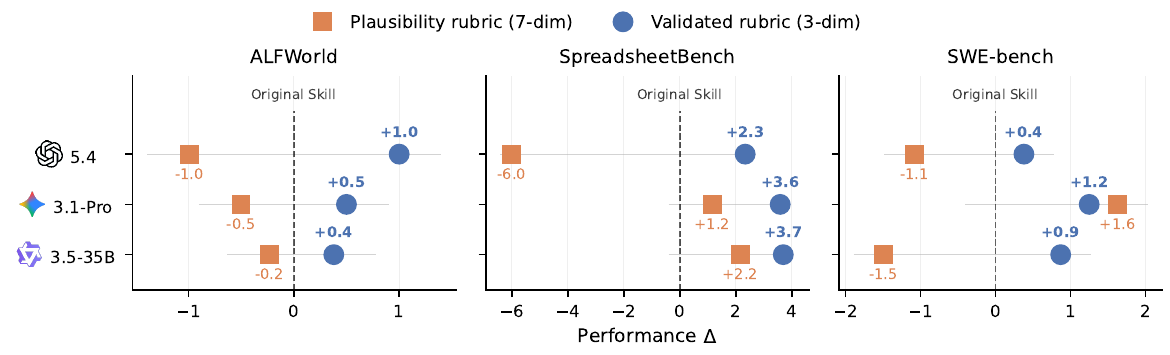}
\caption{\textbf{Effect of meta-skill guidance on downstream skill utility.} The plausibility rubric hurts most times, while the validated rubric improves all the generated skills compared with original skill.}
\label{fig:meta_skill}
\end{figure}

Now we ask whether the Section~\ref{sec:extraction_analysis} finding---that textual plausibility does not predict downstream utility---can be operationalized into actionable criteria that improve both skill evaluation and skill extraction. This is our \textbf{RQ3}: whether our empirical findings can be turned into a concrete, drop-in improvement to skill extraction itself.

A naive starting point is to ask an LLM directly for skill-quality criteria. Doing so yields a generic \textbf{plausibility rubric}: seven dimensions covering clarity, completeness, conciseness, logical structure, formatting, tone, and generality (full list in Appendix~\ref{app:plausibility_rubric}).

\paragraph{Raw and validated rubrics.}
We design a fully automated rubric-discovery pipeline that takes the high-gap skill pairs from the cross-matrix as input. GPT-5.4 first analyzes each pair to extract per-pair differences along which the higher-$\sdelta$ skill outperforms the lower one; these differences are then iteratively merged and consolidated into seven candidate dimensions, which we call the \textbf{raw rubric} (Appendix~\ref{app:contrastive}). We then test which raw dimensions actually predict utility via pairwise evaluation, measuring each dimension's \emph{better-rate}---the proportion of pairs where the higher-$\sdelta$ skill receives more favorable judgments. Three dimensions consistently align with utility: \textbf{Failure Mechanism Encoding}, \textbf{Actionable Specificity}, and \textbf{High-Risk Action Blacklist} (better-rates 64--66\%); together they form the \textbf{validated rubric}.

To verify that the validated rubric carries genuine evaluative signal, we feed it back into the same pairwise-judgment protocol from Section~\ref{sec:extraction_analysis}: on the same 151 high-gap pairs, the judge is now instructed to score each candidate along the three validated dimensions and aggregate them into a single preference. This rubric-guided judgment raises overall judge accuracy from 46.4\% (unguided) to \textbf{73.8\%}. The improvement also extends to the hardest pairs ($\sdeltagap{\geq}5$~pp), where the unguided judge had picked the higher-$\sdelta$ skill only 15.8\% of the time and the guided judge now picks correctly the majority of the time (\Cref{fig:pairwise}). With the validated rubric, the same LLM judge that previously favored more fluent but worse-performing skills becomes a reliable utility predictor.

\paragraph{Meta-skill guided extraction.}
We operationalize the validated rubric as a compact \textbf{meta-skill}: a generation-time prior inserted into the extractor's system prompt.
We compare it against (i) the original (un-guided) extractor prompt and (ii) the same prompt augmented with the plausibility rubric.
As shown in \Cref{fig:meta_skill}, the plausibility rubric \emph{hurts} average performance ($-0.59$~pp), reducing accuracy in 6 of 9 cells, while the validated rubric improves \emph{all nine} cells ($+1.55$~pp average), with the largest gains on SpreadsheetBench ($+2.3$ to $+3.7$~pp).
These results demonstrate the effectiveness of our validated rubric and the resulting meta-skill, which plug directly into any extractor's system prompt without modifying the underlying extraction pipeline.

This closed-loop signal, from diagnostic analysis through dimension validation to measurable downstream improvement, shows that a utility-grounded benchmark can inform not only the evaluation of skills but also the design of skill extraction systems themselves.

\begin{findingbox}
\findinglead{RQ3}We identify a small set of validated rubric dimensions that reliably distinguish higher- from lower-utility skills, and using this rubric as a meta-skill to guide the extractor consistently improves the quality of the generated skills---demonstrating that empirical findings from this study translate directly into a drop-in extraction improvement.
\end{findingbox}

%% file: sections/8_conclusion.tex
\section{Conclusion}
\label{sec:conclusion}

We present a systematic, utility-grounded study of model-generated agent skills across the full lifecycle of experience generation, skill extraction, and skill consumption, spanning five diverse domains and multiple extractors and targets. We find that such skills are beneficial on average but exhibit substantial variance and non-trivial negative transfer, and that neither model scale nor textual plausibility reliably predicts downstream utility. A deep analysis of all three stages, experience generation, skill extraction, and skill consumption, explains where this variance comes from, and we translate these findings into a meta-skill prior, distilled from a validated utility-grounded rubric, which improves extraction in all evaluated cells and plugs directly into any extractor's system prompt. Together, these contributions move agent skill extraction from a heuristic, intuition-driven practice toward a principled, utility-grounded discipline.

%% file: sections/A_appendix.tex
\appendix

\section{Limitations, Future Work, and Broader Impact}
\label{app:limitations}

\paragraph{Limitations and future work.}
Our experimental design intentionally favors interpretability over coverage. We consolidate each target's experience into a single domain-level skill and supply it directly through the system prompt at evaluation time, so that the observed performance change can be attributed as cleanly as possible to the skill itself rather than to retrieval policies, agentic scaffolding, or other confounding components in the pipeline. This minimal setup is what enables the controlled cross-extractor and cross-target comparisons that ground all of our findings. We see two natural directions for future work: scaling to richer agent harnesses (for example, with retrieval, planning, or tool-use scaffolds), and scaling to substantially larger skill libraries containing many fine-grained skills, where additional questions of skill selection, composition, and interference become first-class concerns. We view these directions as complementary to the present study rather than as gaps in it, and as promising avenues for building on the utility-grounded foundation established here.

\paragraph{Broader impact.}
Skill libraries built by language agents are increasingly reused across models and deployments, and our study has two practical implications. On the positive side, the utility-grounded evaluation, the validated rubric, and the resulting meta-skill prior give practitioners a concrete way to screen out skills that look fluent but transfer poorly, reducing the chance of silently shipping skills that degrade performance and saving the compute that would otherwise be spent on unhelpful or harmful skill reuse. On the negative side, more effective skill extraction inherits the general risks of methods that make language agents more capable: skills that raise task success can be repurposed for misuse, and skills extracted from imperfect experience pools may carry over biases or unsafe shortcuts from those traces. Mitigating these risks is part of the future work outlined above, in particular evaluating skill safety in richer agentic harnesses and at larger library scales.

\section{Additional Experimental Details}
\label{app:details}

\subsection{Experience Pool Collection}

For each (target, domain) pair, we run the target model on the training split for multiple rounds, collecting both successful and failed trajectories. Pool sizes vary across domains, reflecting differences in training-split size and per-task cost.

\subsection{Evaluation Details}
\label{app:eval_details}

\paragraph{API access.} For all OpenAI GPT and Google Gemini models in our experiments, we set the reasoning effort (\texttt{reasoning\_effort} for GPT, \texttt{thinking\_level} for Gemini) to \texttt{medium}. GPT models are accessed via the Azure OpenAI API, and Gemini models via the official Google Gemini API.

\paragraph{Data splits.} For each domain $\domain$, we split task instances 1:1 into an experience-generation split $Q^{\text{train}}_\domain$ and a held-out test split $Q^{\text{test}}_\domain$. When an official training split is provided by the benchmark, $Q^{\text{train}}_\domain$ is sampled from it at the same 1:1 proportion relative to $Q^{\text{test}}_\domain$; otherwise we partition the available instances uniformly at random with a fixed seed. The same splits are used across all (extractor, target) combinations within a domain to ensure that observed differences in $\sdelta$ are attributable to the extraction side rather than to evaluation noise.

\paragraph{Repeated runs and aggregation.} All entries in Table~\ref{tab:cross_matrix}, including the \textit{Base} column, are averaged over three independent evaluation runs.

\subsection{Extraction Prompt Templates}
\label{app:prompt_extraction}

This subsection reproduces the prompts that instantiate the two-stage extraction framework of \Cref{sec:extraction_framework}: \textbf{per-trajectory analysis}, which converts each trajectory into a pattern set of success and failure patterns; \textbf{hierarchical consolidation}, which merges pattern sets level by level into a single consolidated pattern set; and a final \textbf{skill synthesis} step, which turns the consolidated pattern set into a schema-conformant skill set via tool calls. Each phase is a single prompted call to $\extractor$.

\paragraph{Per-trajectory analysis prompt.}
For each trajectory, the extractor is prompted to extract up to $K$ success patterns (if the trajectory succeeded) or failure patterns (if it failed). The two cases share the same template (\Cref{tab:prompt_extraction_map}), swapping only the per-type guidance block.

\begin{table}[h]
\centering
\begin{promptbox}
\ttfamily\small
You analyse a single agent trajectory and extract \textnormal{\textit{[success patterns \textbar{} failure patterns]}} -- high-level, reusable, transferable behavioural patterns. Focus on genuinely novel, reusable patterns from THIS trajectory; do NOT try to be exhaustive.\par\medskip
\textbf{What is a pattern?} A pattern is a high-level behaviour that is (1) transferable across a broad class of tasks, (2) actionable enough to follow, (3) non-obvious -- going beyond common sense, and (4) self-contained -- understandable without the original trajectory.\par\medskip
\textbf{Per-type guidance (success).} Capture effective strategies, decision patterns, and methodological insights. Ask: ``What did this agent do RIGHT that other agents facing similar tasks should also do?''\par\medskip
\textbf{Per-type guidance (failure).} Capture error patterns, anti-patterns, and non-obvious pitfalls. Ask: ``What should an agent AVOID doing when facing similar tasks?''\par\medskip
\textbf{Quality requirements.} Each pattern must be (i) high-level and domain-general, (ii) maximally broad in coverage, (iii) information-dense with a concrete description, and (iv) free of task-specific details (no specific file names, identifiers, error messages, or API calls).\par\medskip
\textbf{Constraints.} Extract at most \textnormal{\textit{[K]}} patterns from this trajectory; each as a pattern name and a 2--4 sentence description.\par\medskip
\textbf{Output format.} A JSON list of \{\texttt{type}, \texttt{pattern}, \texttt{description}\} entries plus a brief \texttt{summary}. If no useful patterns are found, return an empty list.
\end{promptbox}
\caption{Per-trajectory analysis prompt: extracts a pattern set from one trajectory.}
\label{tab:prompt_extraction_map}
\end{table}

The accompanying user message contains the trajectory's outcome and reward together with the agent's full step-by-step trace; for interactive environments such as ALFWorld we render compact \texttt{[think]}/\texttt{[action]}/\texttt{[obs]} tuples to avoid header duplication, and for trajectories with a free-form final answer the final answer is appended.

\paragraph{Hierarchical consolidation prompt.}
Pattern sets are merged in groups of $G$ at each level of \Cref{eq:reduce} until a single consolidated pattern set remains. \Cref{tab:prompt_extraction_reduce_inter} shows the merge prompt used at every level.

\begin{table}[h]
\centering
\begin{promptbox}
\ttfamily\small
You receive several pattern sets, each extracted from a different agent trajectory, and merge them into a single consolidated pattern set.\par\medskip
\textbf{Guidelines.}\\
1. \textbf{Deduplicate}: if multiple patterns describe the same or overlapping behaviour, combine them into ONE stronger pattern with the best description.\\
2. \textbf{Generalise}: raise the abstraction level to cover more scenarios; a single well-generalised pattern is worth more than several narrow ones.\\
3. \textbf{Preserve type}: keep success and failure patterns separate; do NOT convert between types.\\
4. \textbf{Preserve quality}: drop vague or low-value patterns; keep concrete, actionable ones.\\
5. \textbf{Prioritise}: when there are too many patterns, retain the most important and broadly applicable ones.\par\medskip
\textbf{Quality requirements.} Each merged pattern must be transferable across tasks, information-dense, non-obvious, and free of task-specific details.\par\medskip
\textbf{Output format.} A JSON object with the consolidated success and failure patterns plus a brief \texttt{summary} of merge decisions.
\end{promptbox}
\caption{Hierarchical consolidation prompt: merges $G$ pattern sets into one.}
\label{tab:prompt_extraction_reduce_inter}
\end{table}

\paragraph{Skill synthesis prompt.}
Once a single consolidated pattern set is obtained, the extractor converts it into the skill set via structured tool-calling operations against a writable store (creation, update, and deletion of skills with schema validation). The system prompt shown in \Cref{tab:prompt_extraction_reduce_final} specifies how patterns are turned into schema-conformant skills.

\begin{table}[h]
\centering
\begin{promptbox}
\ttfamily\small
You receive a consolidated set of success and failure patterns and synthesise them into skills by issuing tool calls against the skill store.\par\medskip
\textbf{Synthesis strategy.}\\
1. \textbf{Integrate both polarities}: a good skill includes both what TO DO (from success patterns) and what to AVOID (from failure patterns).\\
2. \textbf{Organise thematically}: group related patterns into coherent skills around shared themes.\\
3. \textbf{Structure the body clearly}: recommended approaches, common pitfalls, decision criteria for when to apply, and verification methods.\\
4. \textbf{Maintain information density}: every sentence carries actionable content; no platitudes.\\
5. \textbf{Keep the description short}: 1--2 sentences only; all detail goes in the body.\par\medskip
\textbf{Schema requirements.} \texttt{name} (lowercase-hyphen slug, \(\le\) 64 chars); \texttt{description} (1--2 sentences: what class of problems, when to apply); \texttt{body} (Markdown with strategies, pitfalls, decision criteria, verification); optional \texttt{references} and \texttt{scripts}.\par\medskip
\textbf{Budget.} Maximum \textnormal{\textit{[max\_skills]}} skills, each \(\le\) \textnormal{\textit{[max\_skill\_chars]}} characters (strictly enforced); optional total budget \textnormal{\textit{[max\_total\_chars]}}. On a length error, shorten and retry.\par\medskip
\textbf{Operating rules.} Skills MUST be submitted via the creation tool (plain text in the response is ignored). After all skills are added, the extractor signals completion via the finish tool. On tool errors, the issue is fixed and the call is retried.
\end{promptbox}
\caption{Skill synthesis prompt: converts the consolidated pattern set into a schema-conformant skill set via tool calls.}
\label{tab:prompt_extraction_reduce_final}
\end{table}

\paragraph{Optional meta-skill guidance.}
For meta-skill-guided runs (\Cref{sec:meta-skill}), an additional \emph{Extraction Quality Guidance} block --- the validated 3-dimension rubric or the 7-dimension plausibility rubric --- is appended to the per-trajectory and skill-synthesis prompts.

\subsection{Injection Template}
\label{app:injection}

At evaluation time, the extracted skill set is exposed to the target model in one of two ways depending on its size.

\paragraph{Single-skill protocol.}
When there is exactly one skill, we skip the tool protocol and inline the skill body directly into the target's system prompt, using the template in \Cref{tab:prompt_injection}.

\begin{table}[h]
\centering
\begin{promptbox}
\ttfamily\small
\#\#~Skill Reference\par\medskip
Below is a reusable procedural skill extracted from previous successful problem-solving experiences. It may help you solve the current task more effectively. Use it as a reference -- adapt it to the specific task at hand.\par\medskip
\#\#\#~\textnormal{\textit{[skill.name]}}\par\medskip
\textnormal{\textit{[skill.description]}}\par\medskip
\textnormal{\textit{[skill.body]}}\par\medskip
\#\#\#\#~Reference Files \textnormal{(optional)}\\
\textnormal{\textit{[filename]}}: \textnormal{\textit{[content]}} \dots\par\medskip
\#\#\#\#~Script Files \textnormal{(optional)}\\
\textnormal{\textit{[filename]}}: \textnormal{\textit{[content]}} \dots\par\medskip
\textbf{Note:} This skill is an optional aid, not a mandatory procedure. Use your own judgment.
\end{promptbox}
\caption{Single-skill injection prompt template.}
\label{tab:prompt_injection}
\end{table}

\paragraph{Multi-skill protocol.}
When the skill library contains multiple skills, the target consumes them through progressive disclosure: it first calls \texttt{list\_skills} to see names and descriptions, then \texttt{view\_skill} for the full body, and finally \texttt{read\_skill\_file} for any attached references or scripts. For SpreadsheetBench (which runs in a plain-text conversation rather than via OpenAI function calling), these calls are issued as fenced \verb|```skill|~\dots~\verb|```| blocks, analogous to \verb|```python|~\dots~\verb|```| code execution blocks. The corresponding system-prompt section is shown in \Cref{tab:prompt_injection_multi}.

\begin{table}[h]
\centering
\begin{promptbox}
\ttfamily\small
\#\#~Skill Library\par\medskip
You have access to a \textbf{Skill Library} containing \textnormal{\textit{[N]}} reusable procedural skills extracted from previous successful problem-solving experiences. These skills may help you solve the current task more effectively.\par\medskip
\textbf{Available skill tools.}\\
-- \texttt{list\_skills}: see all available skills (name, description).\\
-- \texttt{view\_skill <skill\_name>}: read the full body of a specific skill; also lists attached file names.\\
-- \texttt{read\_skill\_file <skill\_name> <filename>}: read the content of an attached reference or script.\par\medskip
\textbf{How to use skill tools.} Call a skill tool with a \verb|```skill|~\dots~\verb|```| block (NOT a \verb|```python|~\dots~\verb|```| block).\par\medskip
\textbf{Workflow.}\\
1. At the start, call \texttt{list\_skills} to see what is available.\\
2. If a skill seems relevant, call \texttt{view\_skill} to read its body.\\
3. If it has attached files, use \texttt{read\_skill\_file}.\\
4. Adapt the skill's guidance to the specific task.\\
5. After consulting skills, proceed with \verb|```python|~\dots~\verb|```| code blocks.\par\medskip
\textbf{Important notes.} Skill tools are read-only. Each response should contain EITHER a \texttt{```skill```} block OR a \texttt{```python```} block, not both. Skills are optional aids, not mandatory procedures.
\end{promptbox}
\caption{Multi-skill injection prompt template (text-mode skill tool protocol).}
\label{tab:prompt_injection_multi}
\end{table}

\subsection{Extraction Hyperparameters}

All extraction experiments use the mode-based method with the default settings listed in \Cref{tab:extraction_params} unless otherwise noted:

\begin{table}[h]
\centering
\small
\begin{tabular}{ll}
\toprule
Parameter & Value \\
\midrule
Max modes per trajectory & 3 \\
Merge group size & 10 \\
Max skill characters & 3000 \\
Max skills per extraction & 1 \\
Temperature & 1.0 \\
\bottomrule
\end{tabular}
\caption{Default extraction hyperparameters.}
\label{tab:extraction_params}
\end{table}

In the \emph{map} phase, each trajectory is independently analyzed to extract up to 3 behavioral modes (success or failure patterns). In the \emph{reduce} phase, modes are grouped in batches of 10 and iteratively merged into a single consolidated skill set. The extractor model, experience pool, and target model vary across experimental conditions as specified in the main text.

\subsection{Compute Resource}

Closed-source models (GPT-5.4, GPT-5.4-mini, Gemini-3.1-Pro, Gemini-3.1-FL) are accessed through their respective providers' APIs. Open-source models (Qwen3.5-35B, Qwen3.5-9B) are served locally with vLLM~\citep{vllm} on a single node equipped with 8 NVIDIA B200 GPUs, which is sufficient to run all open-source extractors and targets used in the study at the inference scales reported.

\section{Format Normalization Experiment}
\label{app:format}

We test whether skill utility depends on output format by rewriting the strongest extractor's skill on SpreadsheetBench into four canonical formats: \textit{ordered list} (flat numbered steps), \textit{unordered list} (bullet points), \textit{checklist} (checkbox items), and \textit{prose} (flowing paragraphs). Each rewrite is generated by GPT-5.4 with an instruction to preserve all semantic content while converting to the target format; a verification pass confirms content preservation and format compliance. Each format is evaluated on the same test set as in Section~\ref{sec:main_results} for 3 independent rounds.

\paragraph{Statistical test.}
We use the Friedman test, a non-parametric repeated-measures analysis of variance. For each target model, the test treats each task instance as a block and each format (or extractor) as a treatment, testing the null hypothesis that all treatments produce equal performance.
To quantify effect size relative to noise, we compute $\sigma$\text{-ratio} $=\sigma_{\text{factor}}/\sigma_{\text{round}}$, where $\sigma_{\text{factor}}$ is the standard deviation of mean performance across factor levels (formats or extractors) and $\sigma_{\text{round}}$ is the standard deviation across independent evaluation rounds with the same factor level. A $\sigma$-ratio ${>}1$ indicates the factor effect exceeds run-to-run sampling noise.

\begin{table}[h]
\centering
\small
\begin{tabular}{lcccc}
\toprule
 & \multicolumn{2}{c}{Format} & \multicolumn{2}{c}{Extractor} \\
\cmidrule(lr){2-3} \cmidrule(lr){4-5}
Target & $p$ & $\sigma$-r & $p$ & $\sigma$-r \\
\midrule
GPT-5.4       & .983 & 0.11 & $<$.001 & \textbf{1.98} \\
GPT-5.2       & .436 & 0.77 & $<$.001 & \textbf{4.53} \\
GPT-5.4-mini  & .916 & 0.15 & $<$.001 & \textbf{1.31} \\
GPT-5.4-nano  & .881 & 0.40 & .003    & \textbf{1.51} \\
Qwen3.5-35B   & .345 & 0.67 & $<$.001 & \textbf{2.54} \\
Qwen3.5-9B    & .452 & 0.53 & .515    & 0.83 \\
\bottomrule
\end{tabular}
\caption{Format vs.\ extractor effect on SpreadsheetBench. $\sigma$-ratio $=\sigma_{\text{factor}}/\sigma_{\text{round}}$; values ${>}1$ indicate the factor exceeds noise.}
\label{tab:format_norm}
\end{table}

Format has no detectable effect on any target (all $p > 0.34$, all $\sigma$-ratios below 1). In contrast, the extractor control yields significant effects for 5/6 targets ($p < 0.005$) with $\sigma$-ratios well above 1.

\section{Behavioral impact analysis}
\label{app:behavior}

To better understand why the same intervention helps some models but hurts others, we take a closer look at model behavior on SpreadsheetBench. We focus on two representative cases: GPT-5.4, which improves clearly after consuming skills, and Qwen3.5-9B, which regresses under some consumed skills. We describe the behavior changes from three angles: \textbf{decision-making behavior}, \textbf{exploratory behavior}, and \textbf{tool-use behavior}.

\paragraph{Decision-making behavior.}
The main change in decision-making is that skill consumption changes how the model frames the task at the beginning. For GPT-5.4, the consumed skill often moves the model away from writing spreadsheet formulas as the final answer and toward computing the result in Python and writing back the final value. This is especially helpful for cell-level tasks, where formula-based answers may look reasonable but are not always stable under evaluation. In other words, skill consumption mostly works as a strategy correction for GPT-5.4: it does not give the model a new ability, but makes it choose a more reliable solution more often.

For Qwen3.5-9B, the shift is less consistently helpful. After consuming a skill, the model is more likely to leave simple dataframe-style heuristics and follow a workbook-native workflow. This can improve structural correctness, especially on sheet-level tasks, because the model is less likely to overwrite the workbook in a crude way. But this also makes the solution process more complex. On fine-grained tasks, the model more easily makes execution mistakes, so the gain in structure sometimes comes with a drop in robustness.

\paragraph{Exploratory behavior.}
Skill consumption also changes what the model does before editing the workbook. In both models, we more often see early inspection of sheet structure, headers, used ranges, anchors, and target areas. So the effect is not just on the final action; it also changes how the model builds understanding of the workbook first.

For GPT-5.4, this change is usually small but useful. The model becomes a bit less likely to rely on guesses about layout and a bit more likely to ground its edits in the actual workbook structure. For Qwen3.5-9B, the change is stronger. The model more often inspects the workbook before acting, but this extra exploration does not always lead to better execution. In some failure cases, it leads to longer and more complicated reasoning, while the final result is still wrong.

\paragraph{Tool-use behavior.}
The clearest change in tool use is not that models start making new explicit skill calls. Instead, the consumed skill is usually absorbed into the prompt and changes how the existing tools are used.

For GPT-5.4, the toolset itself stays mostly the same, but the usage becomes more grounded. We more often see bounded write-back, anchor-based addressing, and simple checks after writing. For Qwen3.5-9B, the change is larger. The model shifts from \texttt{pandas}-style round-trip rewriting to more \texttt{openpyxl}-based in-place editing. This helps preserve workbook structure, but it also creates more chances to fail when the model cannot reliably carry out the more complex workflow.

\section{Pairwise Skill Evaluation}
\label{app:pairwise}

The unguided pairwise evaluation (Section~\ref{sec:extraction_analysis}) uses GPT-5.4 and 9 independent votes per pair (majority vote). The skill presentation order is randomized per pair to mitigate position bias. The full judge prompt is shown in \Cref{tab:prompt_pairwise}.

\begin{table}[h]
\centering
\begin{promptbox}
\ttfamily\small
You are comparing two agent skill documents
meant to help an AI agent.\\
Domain: \textnormal{\textit{[domain description]}}\par\medskip
Skill 1:\\
\textasciigrave\textasciigrave\textasciigrave\\
\textnormal{\textit{[skill text]}}\\
\textasciigrave\textasciigrave\textasciigrave\par\medskip
Skill 2:\\
\textasciigrave\textasciigrave\textasciigrave\\
\textnormal{\textit{[skill text]}}\\
\textasciigrave\textasciigrave\textasciigrave\par\medskip
Which skill document will produce better agent
performance? You MUST choose one.\\
Reply with JSON: \{"choice": "Skill 1" or "Skill 2"\}
\end{promptbox}
\caption{Unguided pairwise judge prompt template.}
\label{tab:prompt_pairwise}
\end{table}

The domain description provides a one-sentence characterization of the task environment (e.g., ``SpreadsheetBench: the agent writes Python code to manipulate Excel files and produce correct values in specified output cells''). No evaluation rubric or quality criteria are provided, forcing the judge to rely on its own implicit notion of skill quality.

We construct 151 within-group pairs by enumerating all extractor pairs that share the same (target, domain) and whose $|\sdelta|$ gap exceeds 0.5~pp. This threshold excludes near-ties where the ground-truth ranking is unreliable due to evaluation noise.

\section{Alternative Harness Evaluation}
\label{app:harness}

To verify that our SpreadsheetBench results are not artifacts of the Python-script evaluation harness used in the main experiments, we re-evaluate a subset of conditions using two alternative agentic harnesses: Claude Code (CC) and Codex. These harnesses execute spreadsheet tasks via interactive tool-use rather than a fixed script, providing an independent check on skill utility under different execution environments. \Cref{tab:harness} reports the resulting $\sdelta$ matrix.

\begin{table}[h]
\centering
\small
\setlength{\tabcolsep}{4pt}
\begin{tabular}{lcrrrrr|r}
\toprule
Target & Base & GPT-5.4 & GPT-5.4-mini & Qwen3.5-35B & $\TE$ \\
\midrule
CC Opus 4.6    & 70.0 & \textcolor{green!50!black}{+1.0} & \textcolor{green!50!black}{+1.0} & \textcolor{red!70!black}{$-$3.5} & $-$0.5 \\
CC Sonnet 4.6  & 66.5 & \textcolor{green!50!black}{+4.0} & \textcolor{green!50!black}{+1.0} & \textcolor{green!50!black}{+2.0} & +2.3 \\
Codex GPT-5.4      & 53.5 & \textcolor{green!50!black}{+4.0} & \textcolor{red!70!black}{$-$5.5} & \textcolor{green!50!black}{+3.5} & +0.7 \\
Codex GPT-5.4-mini & 42.5 & \textcolor{red!70!black}{$-$1.0} & \textcolor{green!50!black}{+0.0} & \textcolor{red!70!black}{$-$2.0} & $-$1.0 \\
\cmidrule{1-6}
\rowcolor{gray!5} $\EE$ & & \textbf{+2.0} & $-$0.9 & +0.0 & \\
\bottomrule
\end{tabular}
\caption{SpreadsheetBench $\sdelta$ (pp) with alternative agentic harnesses (Claude Code / Codex). \textcolor{green!50!black}{Green}: $\sdelta > 0$; \textcolor{red!70!black}{Red}: $\sdelta < 0$.}
\label{tab:harness}
\end{table}

The overall pattern is consistent with the main results: skill injection yields modest positive gains on average ($\overline{\sdelta}{=}+0.4$~pp), with substantial variance across targets. Notably, stronger targets (CC Opus, Codex GPT-5.4) show positive transfer from GPT-5.4-extracted skills, while the weakest target (Codex GPT-5.4-mini) shows no benefit, echoing the consumption-ability gradient observed in the main experiments.

\section{Meta-Skill Guidance}
\label{app:meta_skill_full}

\Cref{tab:meta_skill} reports the full per-cell accuracy numbers underlying \Cref{fig:meta_skill} in Section~\ref{sec:meta-skill}, including the no-skill baseline and the original (un-guided) skill condition.

\begin{table}[h]
\centering
\small
\setlength{\tabcolsep}{3.5pt}
\begin{tabular}{ll cccc cc}
\toprule
& & & & \multicolumn{2}{c}{\textbf{Guidance}} & \multicolumn{2}{c}{$\boldsymbol{\Delta}$ \textbf{vs Original}} \\
\cmidrule(lr){5-6} \cmidrule(lr){7-8}
\textbf{Domain} & \textbf{Target} & \textbf{No Skill} & \textbf{Original} & \textbf{\shortstack{Plausibility\\(7-dim)}} & \textbf{\shortstack{Validated\\(3-dim)}} & \textbf{\shortstack{Plaus.}} & \textbf{\shortstack{Valid.}} \\
\midrule
\multirow{3}{*}{ALFWorld}
& GPT-5.4      & 68.66 & 75.12 & 74.13 & \textbf{76.12} & \worse{$-$0.99} & \better{+1.00} \\
& Gemini       & 87.56 & 88.31 & 87.81 & \textbf{88.81} & \worse{$-$0.50} & \better{+0.50} \\
& Qwen3.5-35B  & 57.21 & 53.73 & 53.50 & \textbf{54.11} & \worse{$-$0.23} & \better{+0.38} \\
\midrule
\multirow{3}{*}{\shortstack[l]{Spreadsheet-\\Bench}}
& GPT-5.4      & 37.17 & 46.17 & 40.17 & \textbf{48.50} & \worse{$-$6.00} & \better{+2.33} \\
& Gemini       & 37.50 & 33.17 & 34.33 & \textbf{36.75} & \better{+1.16} & \better{+3.58} \\
& Qwen3.5-35B  & 23.83 & 29.33 & 31.49 & \textbf{33.02} & \better{+2.16} & \better{+3.69} \\
\midrule
\multirow{3}{*}{SWE-bench}
& GPT-5.4      & 68.40 & 69.72 & 68.64 & \textbf{70.10} & \worse{$-$1.08} & \better{+0.38} \\
& Gemini       & 66.53 & 69.33 & 70.96 & \textbf{70.58} & \better{+1.63} & \better{+1.25} \\
& Qwen3.5-35B  & 52.92 & 55.00 & 53.51 & \textbf{55.87} & \worse{$-$1.49} & \better{+0.87} \\
\midrule
\multicolumn{2}{l}{\textbf{Average (9 cells)}} & & & & & \worse{$-$0.59} & \better{+1.55} \\
\bottomrule
\end{tabular}
\caption{\textbf{Effect of meta-skill guidance on downstream skill utility} (accuracy \%). The plausibility-based rubric (all 7 dimensions, unscreened) hurts on average; the utility-validated rubric (3 screened dimensions) improves all nine cells.}
\label{tab:meta_skill}
\end{table}

\section{Contrastive Skill Analysis}
\label{app:contrastive}

\subsection{Plausibility Rubric (Naive Baseline)}
\label{app:plausibility_rubric}

The plausibility rubric is obtained by directly asking GPT-5.4 to enumerate seven quality criteria for agent skills, with no exposure to actual skill pairs or downstream-utility data. By construction, the resulting dimensions describe what an LLM \emph{believes} would distinguish a good skill from a bad one---surface qualities of the text---rather than properties grounded in observed utility. \Cref{tab:plausibility_rubric} lists the seven dimensions used as the unguided baseline in Section~\ref{sec:meta-skill}.

\begin{table}[h]
\centering
\small
\begin{tabular}{cl}
\toprule
\# & Dimension \\
\midrule
1 & \textbf{Clarity}: language is unambiguous and free of confusing phrasing \\
2 & \textbf{Completeness}: covers the full scope of typical task scenarios \\
3 & \textbf{Conciseness}: dense and free of unnecessary verbosity \\
4 & \textbf{Logical Structure}: sections are organized in a coherent order \\
5 & \textbf{Formatting Quality}: uses headers, lists, and code blocks where appropriate \\
6 & \textbf{Tone Neutrality}: phrasing is professional and free of biased language \\
7 & \textbf{Generality}: applies broadly across tasks rather than to a single example \\
\bottomrule
\end{tabular}
\caption{Seven dimensions of the \textbf{plausibility rubric}, generated directly by GPT-5.4 without any pair-level or utility-level grounding.}
\label{tab:plausibility_rubric}
\end{table}

\subsection{Raw Rubric (from Contrastive Pipeline)}

The contrastive analysis (Section~\ref{sec:meta-skill}) synthesized seven candidate quality dimensions from recurring themes across 17 high-gap skill pairs. \Cref{tab:rubric_dims} lists all seven dimensions of this \textbf{raw rubric} with their definitions and per-dimension better-rates (the proportion of pairs where the higher-$\sdelta$ skill receives more favorable judgments on that dimension).

\begin{table}[h]
\centering
\small
\begin{tabular}{clc}
\toprule
\# & Dimension & Better-rate \\
\midrule
\textbf{1} & \textbf{Failure Mechanism Encoding}: identifies \emph{why} agents fail, not just \emph{that} they fail & \textbf{65.5\%} \\
\textbf{2} & \textbf{Actionable Specificity}: step-level procedures referencing domain objects/tools & \textbf{66.0\%} \\
3 & Environment/Tool Semantics: encodes how tools and objects actually behave & 63.2\% \\
4 & Strategy Switching Conditions: specifies when to change approach & 47.5\% \\
5 & Boundary Condition Coverage: addresses specific edge cases & 63.0\% \\
\textbf{6} & \textbf{High-Risk Action Blacklist}: forbids specific harmful action patterns & \textbf{64.6\%} \\
7 & Benchmark-Aligned Priorities: addresses what evaluation actually measures & 56.2\% \\
\bottomrule
\end{tabular}
\caption{Seven dimensions of the \textbf{raw rubric}, discovered via the automated contrastive pipeline. Better-rate measures alignment with downstream utility; the three \textbf{bold} dimensions form the \textbf{validated rubric} used for guided evaluation and meta-skill extraction.}
\label{tab:rubric_dims}
\end{table}

\subsection{Representative Contrastive Cases}

\Cref{tab:case_ss,tab:case_alf} show representative best-vs-worst skill pairs from SpreadsheetBench and ALFWorld. We reproduce the full skill text and highlight key passages: \hlgreen{green} marks domain-specific failure mechanisms or executable countermeasures; \hlred{red} marks generic advice that provides little actionable leverage.

\begin{table}[h]
\centering
\scriptsize
\setlength{\tabcolsep}{4pt}
\begin{tabular}{p{0.47\textwidth} p{0.47\textwidth}}
\toprule
\textbf{Higher-$\sdelta$ skill} ($\sdelta{=}+14.7$, ext = Gemini-3.1-FL) & \textbf{Lower-$\sdelta$ skill} ($\sdelta{=}+4.3$, ext = GPT-5.4) \\
\midrule
\hlgreen{Treat spreadsheet files solely as I/O containers. Never rely on the host application to evaluate formulas or perform business logic.}

\textbf{1.\ Proactive Reconnaissance.}
Diagnostic Audit: read all sheets, row counts, headers, sample rows, and merged-cell maps \emph{before} any mutation.
\hlgreen{Dynamic Addressing: search for anchor data (e.g., column headers) to determine indices; never use hardcoded cell references.}
Normalization: establish a cleaning layer before processing.

\textbf{2.\ In-Memory Processing.}
Logic Decoupling: extract data into Python structures; perform all aggregations in memory.
\hlgreen{Avoid Formula Injection: writing formula strings does not trigger calculation engines in headless environments. Always calculate the final static value in Python and write the scalar result.}

\textbf{3.\ Idempotent Write Strategy.}
Atomic Updates: clear target ranges before writing.
\hlgreen{Reverse Iteration: when deleting or rearranging data, iterate bottom-to-top to avoid index-shifting errors.}
Metadata Preservation: use style-preserving libraries.

\textbf{4.\ Post-Execution Validation.}
Verification Loop: perform a post-write audit to confirm output matches expected logic.
Fail-Fast: if an intermediate step fails, simplify rather than patch.

\textbf{Critical Pitfalls:} Formula Injection Fallacy; Verification Blindness; Destructive Mutation; Context-Agnostic Recycling.
&
\textbf{1.} \hlred{Inspect the live artifact first.} Confirm what you are editing and roughly where the relevant scope is before writing anything.

\textbf{2.} \hlred{Resolve the contract before coding.} Determine exact deliverable: edited artifact, formulas vs values, write scope, preservation requirements.

\textbf{3.} Derive logic from semantic anchors. Use headers, labels, markers, nearby formulas; do not rely on fixed coordinates.

\textbf{4.} Normalize into a canonical model. Trim/case-normalize text, parse compound cells, coerce types safely.

\textbf{5.} \hlred{Stage the work.} Separate discovery, computation, mutation, and formatting. Prove the core rule on representative cases before bulk changes.

\textbf{6.} \hlred{Choose the simplest method that matches the contract and runtime.}

\textbf{7.} \hlred{Edit minimally and safely.} Keep changes inside the intended scope and avoid disturbing unrelated parts of the artifact.

\textbf{8.} Round-trip validate the saved result. Reopen the artifact and verify target cells, formulas or values.

\textbf{Pitfalls:} Trusting stale inspection; hardcoding coordinates; guessing ambiguous rules; mixing exploration with mutation; treating successful execution as proof.
\\
\midrule
\multicolumn{2}{p{0.96\textwidth}}{
\textbf{Analysis.}
The higher-$\sdelta$ skill encodes three \emph{domain-specific failure mechanisms} absent from the lower skill:
(1)~the \hlgreen{formula injection fallacy}---formulas are not evaluated in headless execution, so agents must precompute static values;
(2)~\hlgreen{index-shifting errors} during deletion, countered by reverse iteration;
(3)~\hlgreen{dynamic addressing} to avoid hardcoded coordinates.
Each mechanism is paired with an executable remedy.
The lower-$\sdelta$ skill, by contrast, relies on \hlred{process-level directives} (``resolve the contract,'' ``edit minimally'') that are reasonable but too abstract to prevent the concrete failure modes that dominate SpreadsheetBench errors.
} \\
\midrule
\multicolumn{2}{c}{\textbf{Guided rubric judgment (3 validated dimensions)}} \\
\midrule
\multicolumn{2}{c}{\textit{Failure Mechanism Encoding}} \\
\centering \better{\ding{51}} & \centering \worse{\ding{55}} \tabularnewline
\midrule
\multicolumn{2}{c}{\textit{Actionable Specificity}} \\
\centering \better{\ding{51}} & \centering \worse{\ding{55}} \tabularnewline
\midrule
\multicolumn{2}{c}{\textit{High-Risk Action Blacklist}} \\
\centering \better{\ding{51}} & \centering \worse{\ding{55}} \tabularnewline
\bottomrule
\end{tabular}
\caption{Contrastive case: SpreadsheetBench, target = GPT-5.4, $\sdelta$ gap = 10.3~pp.}
\label{tab:case_ss}
\end{table}

\begin{table}[h]
\centering
\scriptsize
\setlength{\tabcolsep}{4pt}
\begin{tabular}{p{0.47\textwidth} p{0.47\textwidth}}
\toprule
\textbf{Higher-$\sdelta$ skill} ($\sdelta{=}+7.5$, ext = Gemini-3.1-Pro) & \textbf{Lower-$\sdelta$ skill} ($\sdelta{=}+1.5$, ext = GPT-5.4) \\
\midrule
\textbf{1.\ Search Strategy \& Spatial Memory.}
Semantic to Systematic: begin searching high-probability locations based on semantics. If not found, transition to an exhaustive sweep of ALL open surfaces and closed receptacles.
\hlgreen{\textbf{Deep Inspection:} never merely observe the exterior of closed receptacles. You MUST explicitly \texttt{open} them and inspect contents to avoid false negatives.}
State and Spatial Memory: maintain a checklist of explored areas to prevent amnesic looping. Memorize incidental item locations for later retrieval.

\textbf{2.\ Strict Pipelining.}
Linear Execution Pipeline: Locate $\to$ Acquire $\to$ Transform $\to$ Navigate $\to$ Deposit. Complete each phase before advancing.
\hlgreen{\textbf{Active State Transformations:} if an object requires a state change (cleaned, heated), locate it, acquire it, transport it to the appliance, invoke the command, and verify.}
Exact Lexical Matching: adhere strictly to the requested target object name; never substitute synonyms.

\textbf{3.\ Preconditions \& Multi-Item Transport.}
\hlgreen{\textbf{Proactive Prerequisite Resolution:} verify and resolve physical preconditions (navigating to proximity, opening destination receptacles) \emph{before} attempting core interactions.}
Incremental Fetch-and-Deliver: for multi-item tasks, use single-item fetch-and-deposit cycles.

\textbf{Pitfalls:} Redundant state verification; semantic fixation; premature goal reversal.
&
\textbf{1.} \hlred{Ground the goal exactly.} Translate the instruction into explicit predicates and act on them in order.

\textbf{2.} \hlred{Find the current bottleneck.} Work backward from success and act on the earliest unmet prerequisite.

\textbf{3.} Search with memory and pivot rules. Start with visible, nearby, semantically likely candidates. Keep a ledger of searched locations, opened objects, confirmed sources, held items, remaining counts. If a location class yields repeated misses, broaden to a new region.

\textbf{4.} \hlred{Manage preconditions through affordances.} Before key actions, make sure access and usability are in place. Treat failed actions as evidence of a missing prerequisite, not a cue to retry.

\textbf{5.} Bank monotonic progress. When you find a valid item, convert it into durable progress quickly. For repeated goals, use acquire-deliver-repeat loops.

\textbf{6.} \hlred{Replan on observation; finish minimally.} After each observation, recheck what is still unsatisfied. Once a valid completion path exists, stop exploring and execute the shortest finish chain.

\textbf{Failure patterns:} searching without coverage memory; shallow inspection treated as proof; stale-plan repetition; endgame thrashing.
\\
\midrule
\multicolumn{2}{p{0.96\textwidth}}{
\textbf{Analysis.}
The higher-$\sdelta$ skill provides three \emph{executable action patterns} tailored to ALFWorld's mechanics:
(1)~\hlgreen{deep inspection}---explicitly \texttt{open} closed containers rather than assuming visibility equals absence;
(2)~\hlgreen{active state transformations}---a concrete locate-acquire-transport-invoke pipeline for state changes;
(3)~\hlgreen{prerequisite resolution}---navigate and open destinations \emph{before} attempting placement.
The lower-$\sdelta$ skill describes the same high-level logic (\hlred{``ground the goal,'' ``find the bottleneck,'' ``manage preconditions''}) but at a level of abstraction that does not map onto ALFWorld's action vocabulary, leaving the agent to rediscover the operational details on its own.
} \\
\midrule
\multicolumn{2}{c}{\textbf{Guided rubric judgment (3 validated dimensions)}} \\
\midrule
\multicolumn{2}{c}{\textit{Failure Mechanism Encoding}} \\
\centering \better{\ding{51}} & \centering \better{\ding{51}} \tabularnewline
\midrule
\multicolumn{2}{c}{\textit{Actionable Specificity}} \\
\centering \better{\ding{51}} & \centering \better{\ding{51}} \tabularnewline
\midrule
\multicolumn{2}{c}{\textit{High-Risk Action Blacklist}} \\
\centering \better{\ding{51}} & \centering \worse{\ding{55}} \tabularnewline
\bottomrule
\end{tabular}
\caption{Contrastive case: ALFWorld, target = GPT-5.4, $\sdelta$ gap = 6.0~pp.}
\label{tab:case_alf}
\end{table}